\documentclass{article}

\usepackage[preprint]{neurips_2026}
\usepackage[utf8]{inputenc}
\usepackage[T1]{fontenc}

\usepackage{hyperref}
\usepackage{url}
\usepackage{microtype}
\usepackage{xspace}

\usepackage{amsfonts}
\usepackage{amsmath}
\usepackage{amssymb}
\usepackage{mathtools}
\usepackage{amsthm}
\usepackage{nicefrac}

\usepackage[table,dvipsnames]{xcolor}
\usepackage{graphicx}
\usepackage{subcaption}
\usepackage{caption}
\usepackage{booktabs}
\usepackage{multirow}
\usepackage{makecell}
\usepackage{array}
\usepackage{tabularx}
\usepackage{adjustbox}
\usepackage{longtable}
\usepackage{rotating}
\usepackage{lscape}
\usepackage{colortbl}

\usepackage{algorithm}
\usepackage{algpseudocode}

\usepackage{enumitem}
\usepackage{siunitx}
\usepackage{arydshln}
\usepackage[capitalize,noabbrev]{cleveref}

\usepackage{amsmath,amsfonts,bm}

\def\eqref#1{equation~\ref{#1}}
\def\1{\bm{1}}

\def\vmu{{\bm{\mu}}}

\def\vb{{\bm{b}}}
\def\vc{{\bm{c}}}
\def\vd{{\bm{d}}}

\def\vv{{\bm{v}}}

\def\vx{{\bm{x}}}

\def\mW{{\bm{W}}}

\DeclareMathAlphabet{\mathsfit}{\encodingdefault}{\sfdefault}{m}{sl}
\SetMathAlphabet{\mathsfit}{bold}{\encodingdefault}{\sfdefault}{bx}{n}

\def\sN{{\mathbb{N}}}

\def\sR{{\mathbb{R}}}
\def\sS{{\mathbb{S}}}

\newcolumntype{L}[1]{>{\raggedright\arraybackslash}p{#1}}
\newcolumntype{Y}{>{\raggedright\arraybackslash}X}
\newcolumntype{C}{>{\centering\arraybackslash}m{1.1cm}}

\definecolor{gainGreen}{RGB}{0,115,70}

\definecolor{tableheader}{RGB}{242,246,248}
\definecolor{tablesubheader}{RGB}{249,251,252}
\definecolor{gainGreen}{RGB}{0,100,60}
\newcommand{\ours}{VS2\xspace}
\newcommand{\oursplus}{VS2$^{++}$\xspace}

\newcommand{\posd}[1]{\textcolor{gainGreen}{#1}}
\newcommand{\negd}[1]{\textcolor{BrickRed}{#1}}
\newcommand{\gain}[1]{\textcolor{gainGreen}{\footnotesize$\uparrow$#1}}
\newcommand{\loss}[1]{\textcolor{BrickRed}{#1}}
\newcommand{\accinline}[2]{#1\,(\,#2\,)}

\theoremstyle{plain}
\newtheorem{proposition}{Proposition}[section]

\theoremstyle{definition}

\theoremstyle{remark}
\newtheorem{remark}{Remark}[section]

\title{Beyond Interpretability: When, Why, and How Sparse Autoencoders Enable Label-Free Visual Steering}

\author{%
  \makebox[\textwidth][c]{%
    Gerasimos Chatzoudis$^{1}$\thanks{Corresponding author.}\hspace{1.5em}%
    Zhuowei Li$^{1}$\hspace{1.5em}%
    Gemma E. Moran$^{2}$\hspace{1.5em}%
    Hao Wang$^{1}$\hspace{1.5em}%
    Dimitris N. Metaxas$^{1}$%
  } \\[4pt]
  $^{1}$Department of Computer Science, Rutgers University \\
  $^{2}$Department of Statistics, Rutgers University \\[2pt]
  \texttt{\{gc745, zl502, hw488, dnm\}@cs.rutgers.edu} \hspace{1em}
  \texttt{gm845@stat.rutgers.edu}
}

\begin{document}

\maketitle

Sparse Autoencoders (SAEs) are increasingly used to interpret foundation models, but their role as an actionable intervention space remains less understood, especially in vision. We study whether sparse visual features can be used not only for post-hoc analysis, but also to steer frozen vision-language models. We introduce \textbf{Visual Sparse Steering (\ours)}, a label-free method that trains a top-$k$ SAE on unlabeled activations from a frozen CLIP image encoder and, at test time, constructs an interpretable steering vector by amplifying the input's active sparse features and decoding the induced change. We show that this procedure admits a closed-form decomposition as \emph{centroid-deviation steering}: each input is moved along its deviation from the SAE-learned centroid. The residual term is controlled exactly by the SAE's per-sample reconstruction error, measured by FVU, yielding an FVU-based residual bound and motivating a reliability gate that falls back to zero-shot CLIP when SAE reconstruction is unreliable. With target-domain SAEs trained on unlabeled CLIP image-encoder activations, \ours improves zero-shot accuracy across nine image-classification datasets, achieving gains up to $+4.12\%$ with less than $0.1\%$ additional inference compute. Finally, a controlled upper-bound study, \oursplus, shows that selective amplification of sparse features can yield gains up to $+21.44\%$, exposing a \emph{reconstruction-vs-task saliency gap}: features salient for reconstruction need not align with features useful for downstream prediction.

\section{Introduction}
\label{sec:intro}

Sparse Autoencoders (SAEs) have become a central tool in mechanistic interpretability, where they decompose model activations into sparse, human-interpretable features~\citep{bricken2023monosemanticity, gao2024scaling}. Most work treats SAE latents as objects of analysis: they help explain what a model represents and which features activate on a given input. A key open question is whether these sparse features can move beyond post-hoc interpretation and serve as an actionable intervention space for downstream tasks.

This question is particularly important in vision-language models. Adapting frozen models such as CLIP often requires labeled data, prompt tuning, test-time optimization, or carefully constructed contrastive examples. Test-time adaptation methods can be effective, but typically rely on backpropagation through large encoders~\citep{tpt, difftpt, ctpt}; steering vectors are cheaper, but usually require positive and negative anchors or other contrastive supervision~\citep{turner2023activation, zou2023representation, liu2024incontext}. In contrast, SAE features are learned from unlabeled activations and expose interpretable sparse directions in representation space. However, it remains unclear when such features are effective, reliable, or task-relevant for downstream vision tasks.

We study whether SAE sparse features can be used for label-free downstream steering. We focus on zero-shot image classification with frozen CLIP models and introduce \textbf{Visual Sparse Steering (\ours)}, which trains a top-$k$ SAE on unlabeled image activations and uses its active sparse features to construct an instance-specific, test-time steering vector. Given an input image, \ours amplifies its active SAE features and decodes the induced change, producing a direction aligned with the SAE's most salient reconstruction features for that input. This requires no labels, contrastive anchors, CLIP weight updates, or test-time backpropagation.

\begin{figure}[t]  
  \centering
  \begin{subfigure}[b]{0.495\linewidth}
    \centering
    \includegraphics[width=\linewidth]{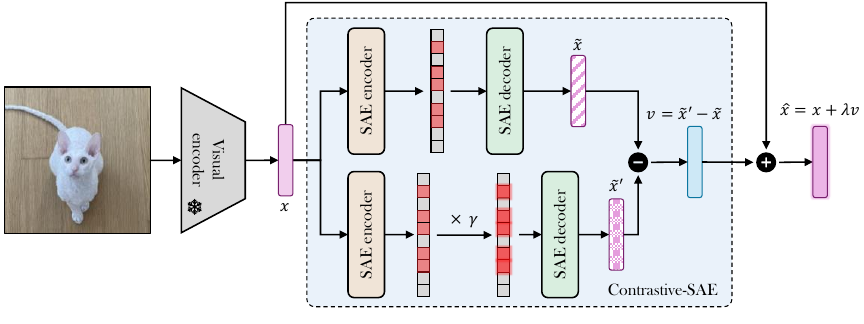}
    \caption{\ours: Visual Sparse Steering w/o Contrastive Data}
    \label{fig:ours}
  \end{subfigure}
  \hfill
  \begin{subfigure}[b]{0.495\linewidth}
    \centering
    \includegraphics[width=\linewidth]{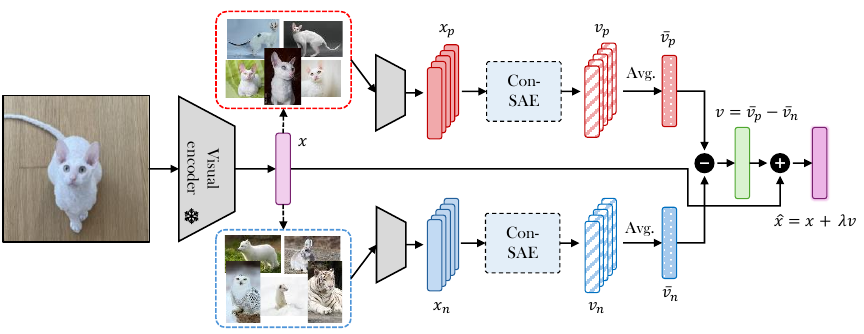} 
    \caption{\oursplus: Selective Visual Sparse Steering}
    \label{fig:ours-plus}
  \end{subfigure}
  \caption{Overview of \ours and \oursplus: \ours constructs a steering vector by uniformly amplifying active sparse features, while \oursplus selectively amplifies sparse features given external corpus.}
  \label{fig:method-diagram}
\end{figure}

Across nine datasets and two CLIP backbones, \ours consistently improves zero-shot classification and matches or exceeds prompt-based test-time adaptation baselines while adding less than $0.1\%$ inference compute. Beyond empirical gains, we show that \ours admits an interpretable closed-form decomposition as \emph{centroid-deviation steering}: each embedding is moved away from the SAE-learned population centroid, suppressing shared structure and amplifying what makes the image distinctive. This provides a geometric explanation for why a label-free intervention can improve classification.

A central challenge is reliability. When the SAE poorly reconstructs the target distribution, sparse features can become unreliable and steering can harm performance. We formalize this failure mode through fraction of variance unexplained (FVU), showing that the residual magnitude in the steering direction is controlled by the SAE's per-sample reconstruction error. This motivates a label-free FVU-gated fallback that disables steering and reverts to zero-shot CLIP when reconstruction is unreliable.

Finally, we identify a deeper limitation of using SAEs for downstream tasks: features salient for reconstruction need not be the features most useful for prediction. We call this the \emph{reconstruction-vs-task saliency gap}. To quantify the headroom from selecting task-relevant features, we introduce \oursplus as a controlled selective-amplification study using retrieved neighbors and CLIP pseudo-labels. The gap between \ours and oracle \oursplus shows that SAE-based downstream steering is limited less by the absence of useful sparse features than by the difficulty of selecting the task-relevant subset reliably.

\noindent\textbf{Contributions.}
We make five contributions: (i) we introduce \ours, a label-free SAE-based steering method for frozen vision-language models; (ii) we show consistent zero-shot classification gains across nine datasets and two CLIP backbones with negligible inference overhead; (iii) we derive a closed-form centroid-deviation interpretation with an FVU-based faithfulness guarantee; (iv) we use FVU as a per-sample reliability gate under distribution shift; and (v) we identify the reconstruction-vs-task saliency gap through \oursplus and prototype-aligned SAE training.

\section{Related Work}
\label{sec:related}
\textbf{Mechanistic Interpretability and Sparse Autoencoders.}
Several traditional approaches exist for interpretability in vision models, including feature visualization \citep{simonyan2014visualising, zeiler2014visualizing, olah2017feature} and network dissection \citep{bau2017network, oikarinen2022clip}. Mechanistic interpretability seeks to systematically analyze and understand neural networks \citep{elhage2021mathematical, olah2020zoom}, but it faces challenges due to polysemantic neurons i.e. units that activate in response to multiple, seemingly unrelated inputs \citep{elhage2022toy}. This phenomenon arises from superposition, where networks encode more features than the available dimensions allow, forcing different concepts to share the same activations~\citep{elhage2022toy}. Sparse Autoencoders (SAEs) have been explored to mitigate superposition by applying sparse dictionary learning to model internals~\citep{bricken2023monosemanticity}. Recent efforts have leveraged SAEs to uncover interpretable features within LLMs ~\citep{templeton2024scaling, cunningham2023sparse, gao2024scaling}. ~\citet{joshi2025identifiable} train SAEs on embedding differences to disentangle multiple concept shifts, enabling precise interventions in model activations without requiring direct supervision. While these methods focus on language models, our work extends sparse steering to the vision domain, where applications of SAEs to \textbf{vision models} remain \textit{comparatively} underexplored. 

While SAEs have been explored for feature analysis, generative modeling, and concept disentanglement in the visual domain~\citep{ bhalla2024interpreting, stevens2025sparse, surkov2024one, fel2025archetypal, thasarathan2025universal}, these works focus primarily on interpretability rather than downstream performance. In contrast, our method leverages SAEs to \textit{actively steer} vision models in a label-free, test-time setting to improve classification. Related efforts like~\cite{joseph2025steering} study CLIP’s steerability on typographic attacks, while Patch-SAE~\citep{lim2024sparse} improves classification accuracy via \emph{class-conditioned} latent masking. Our approach, on the other hand, requires no class labels and avoids class-based activation aggregation during training, enabling broader applicability without reliance on external supervision or gradient updates.

% \vspace{-0.1cm}
\textbf{Steering vectors.} Steering Vector (SV) methods~\citep{turner2023activation, park2023the, hernandez2024linearity, NIPS2013_9aa42b31}, also known as representation engineering~\citep{zou2023representation}, construct a directional task vector and apply it in the latent space to change the target model's behavior at inference time.
In LLMs/MLLMs, SVs are used to enhance security~\citep{liu2024incontext}, truthfulness~\citep{NEURIPS2023_81b83900}, reduce hallucinations~\citep{vista}, and improve efficiency~\citep{i2cl}.
Interestingly, prior work has shown that in VLMs, \emph{visual cues are influenced by language}, and that biases in the model’s response can be steered through simple natural language prompts~\citep{gavrikov2025can}.
In contrast, we focus on steering latent representations in vision models without any language input. 
Recent work has demonstrated that sparse representations can improve interpretability and disentanglement in steering directions~\citep{bayat2025steering, makelov2024sparse}.
Unlike these methods, which rely on supervised contrastive examples or training data, our approach discovers meaningful sparse directions in vision models without requiring labeled positive/negative concept pairs, making it more adaptable to general visual representations.

\textbf{Test-Time Adaptation.} Recent advances in improving the generalization of vision-language models have introduced a
variety of prompt-tuning approaches, such as CoOp~\citep{coop}, CoCoOp~\citep{cocoop}, and
MaPLe~\citep{khattak2023maple}, along with adapter-based techniques like
Tip-Adapter~\citep{zhang2021tip} and CLIP-Adapter~\citep{gao2024clip}. Despite their effectiveness,
these methods typically assume access to a small number of labeled target-domain examples (often
in a few-shot setting).
In contrast, Test-Time Adaptation methods aim to improve robustness under distribution shift by
adapting a pre-trained model using only unlabeled test samples at inference time~\citep{liu2021ttt++,
sun2020test, wang2020tent, gao2022visual}. 
Specifically, Test-Time Prompt Tuning (TPT)~\citep{tpt} introduced entropy-minimizing prompt optimization over augmented views, inspiring subsequent methods such as DiffTPT~\citep{difftpt}, which leverages diffusion-based augmentations, and C-TPT~\citep{ctpt}, which incorporates calibration-aware objectives. Although effective, these strategies typically require optimization loops, multiple augmentations, and backpropagation through large encoders, resulting in substantial computational and memory overhead at inference. In contrast, VS2 performs a single forward pass with sparse SAE-guided steering, avoiding test-time optimization while still providing measurable gains.

\section{Method}
\label{sec:method}

\textbf{Preliminaries (top-$k$ SAE).}
Let $p_{\mathrm{train}}$ denote the empirical distribution over CLS-token activations $\vx\in\sR^d$ of a frozen vision encoder, and let $\vmu:=\mathbb{E}_{p_{\mathrm{train}}}[\vx]$ be its mean. A top-$k$ Sparse Autoencoder (SAE) with latent width $n$ encodes $\vx$ into sparse features $\vc\in\sR^{n}$ and reconstructs $\tilde{\vx}\in\sR^d$ via
\begin{equation}
\begin{aligned}
\vc \,&=\, \mathrm{TopK}_k\!\bigl(\mW_{\mathrm{enc}}(\vx-\vb_{\mathrm{pre}})\bigr),\\[1pt]
\tilde{\vx} \,&=\, D_c(\vc) \,:=\, \mW_{\mathrm{dec}}\,\vc + \vb_{\mathrm{pre}},
\end{aligned}
\label{eq:sae}
\end{equation}
where $\mathrm{TopK}_k(\cdot)$ retains the $k$ entries of largest absolute value and zeros the rest. Following \citet{gao2024scaling}, the SAE is trained to minimize $\|\vx-\tilde{\vx}\|_2^2$ over $p_{\mathrm{train}}$ with $\vb_{\mathrm{pre}}$ initialized to the empirical mean of $\vx$. We write $\vd_i\in\sR^d$ for the $i$-th column of $\mW_{\mathrm{dec}}$ and $\sS_k(\vx):=\{i:c_i\neq 0\}$ for the active set, so $|\sS_k(\vx)|=k$ and $\mW_{\mathrm{dec}}\vc=\sum_{i\in\sS_k(\vx)}c_i\,\vd_i$.

\subsection{Visual Sparse Steering (\ours)}
\label{sec:vs2}

Classical steering-vector methods specify a direction by contrasting curated positive and negative anchor examples~\citep{turner2023activation, zou2023representation}. \ours instead derives a per-instance steering direction directly from the vision encoder's own internal activations: it amplifies the SAE's active sparse features, decodes, and subtracts the un-amplified reconstruction. The construction has an exact closed-form interpretation as centroid-deviation steering and a quantitative faithfulness guarantee tied to the SAE's reconstruction error (\cref{prop:vs2-centroid}). We present \ours as the first systematic study of SAE features as an actionable, label-free control mechanism for vision foundation models, going beyond their established role as interpretability tools.

\textbf{Construction.} Given a frozen vision encoder and an SAE trained on its CLS-token activations, \ours applies three steps for amplification $\gamma>1$ and steering scale $\lambda>0$:
\begin{equation}
\begin{aligned}
\vc \,&=\, \mathrm{TopK}_k\!\bigl(\mW_{\mathrm{enc}}(\vx-\vb_{\mathrm{pre}})\bigr),\\[1pt]
\vv \,&=\, D_c(\gamma\vc) \,-\, D_c(\vc),\\[1pt]
\hat{\vx} \,&=\, \bigl(\vx + \lambda\vv\bigr)\,\tfrac{\|\vx\|_2}{\|\vx + \lambda\vv\|_2}.
\end{aligned}
\label{eq:vs2}
\end{equation}
The first line \emph{encodes} the embedding into its sparse features. The second \emph{steers}: it amplifies the active sparse features by $\gamma$, decodes, and subtracts the un-amplified reconstruction; the difference cancels the SAE's centering bias and isolates the contribution of the active features. The third \emph{rescales} to preserve the embedding norm, so the steered point remains comparable to zero-shot CLIP at classification time. The intervention space is the sparse decoder basis $\{\vd_i\}_{i\in\sS_k(\vx)}$, $k$ directions extracted from the encoder's own activation geometry, many of which empirically align with coherent visual concepts (\cref{appendix:sae}). Because $\sS_k(\vx)$ is input-dependent and the top-$k$ support already identifies features the encoder finds most informative for reconstructing $\vx-\vb_{\mathrm{pre}}$, the steering direction is per-instance and label-free, with no contrastive anchors required. \cref{fig:method-diagram} (left) shows the schematic.

\subsection{A Geometric View of \ours}
\label{sec:vs2-geometry}

We rewrite \eqref{eq:vs2} in closed form. Let $\bm{\epsilon}(\vx) := \tilde{\vx} - \vx$ be the per-sample reconstruction residual and $\mathrm{FVU}(\vx) := \|\bm{\epsilon}(\vx)\|_2^2/\|\vx-\vmu\|_2^2$ the per-sample fraction of variance unexplained.

\begin{proposition}[Centroid-deviation steering with faithfulness guarantee]
\label{prop:vs2-centroid}
For $\gamma>1$, $\lambda>0$, and $\alpha:=\lambda(\gamma{-}1)$,
\begin{equation}
\boxed{\;
\vx + \lambda\vv \,=\, \vx + \alpha(\vx-\vb_{\mathrm{pre}}) + \alpha\,\bm{\epsilon}(\vx),\;}
\label{eq:centroid-steering}
\end{equation}
with residual magnitude $\|\alpha\,\bm{\epsilon}(\vx)\|_2 = \alpha\sqrt{\mathrm{FVU}(\vx)}\,\|\vx-\vmu\|_2$.
\end{proposition}

\textit{Proof}
The decoder is affine, so $\vv = D_c(\gamma\vc) - D_c(\vc) = (\gamma{-}1)\,\mW_{\mathrm{dec}}\vc$ (the bias cancels). Substituting $\mW_{\mathrm{dec}}\vc = (\vx-\vb_{\mathrm{pre}}) + \bm{\epsilon}(\vx)$ from \eqref{eq:sae} and scaling by $\lambda$ gives \eqref{eq:centroid-steering}. The residual magnitude is direct from the definition of $\mathrm{FVU}$.
% \end{proof}

The steered embedding is the input plus an amplification of its deviation from the SAE-learned centroid $\vb_{\mathrm{pre}}\!\approx\!\vmu$, plus a residual of magnitude $\alpha\sqrt{\mathrm{FVU}(\vx)}\,\|\vx-\vmu\|_2$. We highlight three observations.

\begin{itemize}[noitemsep,topsep=2pt,parsep=2pt,leftmargin=1.4em]
\item \textbf{Centroid-deviation steering.} Population-mean structure (common backgrounds, mean texture, dataset-wide directions) is absorbed into $\vb_{\mathrm{pre}}$ and subtracted; what remains amplifies what makes the input distinctive.
\item \textbf{Non-uniform direction from a scalar amplification.} Although $\gamma$ is scalar, $\vv = (\gamma{-}1)\sum_{i\in\sS_k(\vx)}c_i\,\vd_i$ weights each active feature by its own activation; top-$k$ has already filtered features with smaller pre-activations.
\item \textbf{Faithfulness from FVU.} \ours preserves $\vx$ and admits the residual only at scale $\alpha$, in contrast to using the SAE reconstruction $\tilde{\vx}=\vx+\bm{\epsilon}$ directly, which would ingest the full residual. The signal-to-noise ratio $\|\vx-\vb_{\mathrm{pre}}\|_2/\|\bm{\epsilon}(\vx)\|_2 \approx 1/\sqrt{\mathrm{FVU}(\vx)}$ (under faithful centering) makes per-sample FVU the natural reliability gate (\cref{sec:safe}).
\end{itemize}

\begin{remark}[Toward selective amplification]
\label{rem:uniform-to-selective}
\textbf{Replacing the scalar $\gamma$ by a vector $\bm{\gamma}\in\sR^n$} via $\vc'=\bm{\gamma}\odot\vc$ yields $\vv=\sum_{i\in\sS_k(\vx)}(\gamma_i{-}1)\,c_i\,\vd_i$, which spans the full $|\sS_k(\vx)|$-dimensional decoder subspace rather than a single direction. \oursplus realizes a related extension by replacing $\vc$ itself with contrastive sparse features aggregated from retrieved neighbors.
\end{remark}

\subsection{Headroom Analysis via Selective Amplification (\oursplus)}
\label{sec:vs2-plus}

\cref{rem:uniform-to-selective} noted that scalar amplification collapses the steering direction to a single direction within the decoder subspace. To measure the accuracy headroom available to selective, per-feature amplification, we introduce \oursplus as a controlled study: it constructs a contrastive direction from retrieved neighbors using CLIP pseudo-labels (and, as an upper bound, ground-truth labels). \oursplus is not a competing deployment-ready method but an instrument for quantifying this headroom.

\textbf{Pseudo-labeled neighbor groups.} Given a query embedding $\vx_q$ and an unlabeled corpus $\{\vx_i\}_{i=1}^M$, retrieve the $N$ nearest neighbors $\sN(\vx_q)$ in cosine similarity and assign each a CLIP pseudo-label $\hat y_i := \arg\max_y P(y\mid\vx_i)$. Let $\hat y_q := \arg\max_y P(y\mid\vx_q)$ be the query's own pseudo-label, and partition $\sN(\vx_q)$ by pseudo-label agreement with $\vx_q$:
\begin{equation}
\sS^+ \,:=\, \{\vx_i \in \sN(\vx_q): \hat y_i = \hat y_q\},
\quad
\sS^- \,:=\, \sN(\vx_q)\setminus\sS^+.
\label{eq:vs2-plus-groups}
\end{equation}

\textbf{Contrastive steering direction.} For each $\vx_i\in\sN(\vx_q)$, let $\vc_i$ denote its sparse features and $\vv_i$ its \ours direction. Define the group-averaged sparse features $\bar{\vc}^{\pm}:=\tfrac{1}{|\sS^{\pm}|}\sum_{\vx_i\in\sS^{\pm}}\vc_i\in\sR^{n}$. The \oursplus contrastive steering vector is
\begin{equation}
\begin{aligned}
\bar{\vv} \,&=\, \tfrac{1}{|\sS^+|}\!\!\sum_{\vx_i\in\sS^+}\!\!\vv_i
\,-\,
\tfrac{1}{|\sS^-|}\!\!\sum_{\vx_j\in\sS^-}\!\!\vv_j\\[1pt]
&=\, (\gamma{-}1)\,\mW_{\mathrm{dec}}\bigl(\bar{\vc}^+ - \bar{\vc}^-\bigr),
\end{aligned}
\label{eq:vs2-plus}
\end{equation}
where the second equality uses $\vv_i=(\gamma{-}1)\mW_{\mathrm{dec}}\vc_i$ from \cref{prop:vs2-centroid} and linearity of $\mW_{\mathrm{dec}}$. The query embedding is steered by replacing $\vv$ with $\bar{\vv}$ in step~(3) of \eqref{eq:vs2}.

\textbf{Selective amplification.} \oursplus thus anchors steering on features consistently activated within $\sS^+$ while suppressing those highlighted by hard negatives from the same local neighborhood. \cref{fig:method-diagram} (right) shows the schematic. The gap between \ours's uniform amplification and \oursplus's selective amplification quantifies the \emph{reconstruction-vs-task saliency} gap, which we identify as the central open problem for SAE-based downstream methods.

\section{Experiments}
\label{sec:experiments}

\begin{table}[t]
\centering
\caption{\textbf{Benchmarking Visual Sparse Steering.}
Left: zero-shot top-1 accuracy (\%) with and without retrieval on CIFAR-100, CUB-200, and Tiny-ImageNet using ViT-B/32 and ViT-B/16. Right: TTA comparison and inference-time compute for ViT-B/32.}
\label{tab:acc_delta_inline}
\scriptsize
\setlength{\tabcolsep}{2.5pt}
\renewcommand{\arraystretch}{0.86}

\begin{minipage}[t]{0.69\linewidth}
\centering
\vspace{0pt}
\textbf{(a) Main accuracy results}
\vspace{2pt}

\begin{adjustbox}{width=\linewidth}
\begin{tabular}{l cc cc cc}
\toprule
& \multicolumn{2}{c}{\textbf{CIFAR-100}}
& \multicolumn{2}{c}{\textbf{CUB-200}}
& \multicolumn{2}{c}{\textbf{Tiny-IN}} \\
\cmidrule(lr){2-3}\cmidrule(lr){4-5}\cmidrule(lr){6-7}
\textbf{Method}
& ViT-B/32 & ViT-B/16
& ViT-B/32 & ViT-B/16
& ViT-B/32 & ViT-B/16 \\
\midrule

\multicolumn{7}{c}{\textbf{Label-free, unsupervised steering}}\\[2pt]
$\text{CLIP}_{\text{ZS}}$ &
  \accinline{61.07}{0} & \accinline{63.96}{0} &
  \accinline{51.76}{0} & \accinline{55.06}{0} &
  \accinline{56.64}{0} & \accinline{61.08}{0} \\

$\text{SAE}_{\text{REC}}^{\text{A}}$ &
  \accinline{58.01}{\negd{-3.06}} & \accinline{64.05}{\posd{+0.09}} &
  \accinline{47.45}{\negd{-4.31}} & \accinline{51.81}{\negd{-3.25}} &
  \accinline{30.56}{\negd{-26.08}} & \accinline{52.96}{\negd{-8.12}} \\

$\text{SAE}_{\text{REC}}^{\text{F}}$ &
  \accinline{58.22}{\negd{-2.85}} & \accinline{63.42}{\negd{-0.54}} &
  \accinline{48.08}{\negd{-3.68}} & \accinline{51.43}{\negd{-3.63}} &
  \accinline{36.33}{\negd{-20.31}} & \accinline{54.84}{\negd{-6.24}} \\

$\text{SAE}_{\text{REC}}^{\text{F}+\gamma}$ &
  \accinline{62.69}{\posd{+1.62}} & \accinline{66.81}{\posd{+2.85}} &
  \accinline{49.41}{\negd{-2.35}} & \accinline{53.28}{\negd{-1.78}} &
  \accinline{39.49}{\negd{-17.15}} & \accinline{58.81}{\negd{-2.27}} \\

\textbf{\ours} &
  \accinline{\textbf{64.52}}{\posd{\textbf{+3.45}}} &
  \accinline{\textbf{68.08}}{\posd{\textbf{+4.12}}} &
  \accinline{\textbf{52.69}}{\posd{\textbf{+0.93}}} &
  \accinline{\textbf{56.14}}{\posd{\textbf{+1.08}}} &
  \accinline{\textbf{58.14}}{\posd{\textbf{+1.50}}} &
  \accinline{\textbf{62.92}}{\posd{\textbf{+1.84}}} \\

\addlinespace[3pt]
\multicolumn{7}{c}{\textbf{RAG-enhanced, oracle neighbor labels}}\\[2pt]
Weighted RAG &
  \accinline{76.43}{\posd{+15.36}} & \accinline{69.78}{\posd{+5.82}} &
  \accinline{58.32}{\posd{+6.56}} & \accinline{60.65}{\posd{+5.59}} &
  \accinline{72.53}{\posd{+15.89}} & \accinline{67.84}{\posd{+6.75}} \\

CLIP Steering &
  \accinline{81.85}{\posd{+20.78}} & \accinline{84.12}{\posd{+20.16}} &
  \accinline{56.42}{\posd{+4.66}} & \accinline{61.51}{\posd{+6.45}} &
  \accinline{\textbf{80.38}}{\posd{\textbf{+23.74}}} & \accinline{84.07}{\posd{+22.99}} \\

\textbf{\oursplus} &
  \accinline{\textbf{81.95}}{\posd{\textbf{+20.88}}} &
  \accinline{\textbf{85.40}}{\posd{\textbf{+21.44}}} &
  \accinline{\textbf{58.84}}{\posd{\textbf{+7.08}}} &
  \accinline{\textbf{61.91}}{\posd{\textbf{+6.85}}} &
  \accinline{73.27}{\posd{+16.63}} &
  \accinline{\textbf{81.55}}{\posd{\textbf{+20.47}}} \\

\addlinespace[3pt]
\multicolumn{7}{c}{\textbf{RAG-enhanced, CLIP pseudo-labels}}\\[2pt]
CLIP Steering &
  \accinline{\textbf{77.22}}{\posd{\textbf{+16.15}}} &
  \accinline{78.97}{\posd{+15.01}} &
  \accinline{47.89}{\negd{-3.87}} &
  \accinline{53.95}{\negd{-1.11}} &
  \accinline{\textbf{74.06}}{\posd{\textbf{+17.42}}} &
  \accinline{76.79}{\posd{+15.71}} \\

\textbf{\oursplus} &
  \accinline{77.11}{\posd{+16.04}} &
  \accinline{\textbf{79.14}}{\posd{\textbf{+15.18}}} &
  \accinline{\textbf{52.81}}{\posd{\textbf{+1.05}}} &
  \accinline{\textbf{57.02}}{\posd{\textbf{+1.96}}} &
  \accinline{72.12}{\posd{+15.48}} &
  \accinline{\textbf{76.92}}{\posd{\textbf{+15.84}}} \\

\bottomrule
\end{tabular}
\end{adjustbox}
\end{minipage}
\hfill
\begin{minipage}[t]{0.29\linewidth}
\centering
\vspace{0pt}

{\scriptsize\textbf{(b) TTA comparison}\par}
\vspace{1pt}

{\fontsize{5.7}{6.3}\selectfont
\setlength{\tabcolsep}{1.8pt}
\renewcommand{\arraystretch}{0.72}
\begin{tabular}{@{}lccc@{}}
\toprule
Method & CIFAR & CUB & Tiny \\
\midrule
ZS & 61.07 & 51.76 & 56.64 \\
Ens & 63.66 & 51.54 & 61.39 \\
TPT & 64.09 & 51.83 & 62.77 \\
C-TPT & 64.86 & 52.54 & \textbf{63.20} \\
Diff-TPT & 63.04 & \textbf{52.80} & 61.60 \\
\midrule
\textbf{\ours} & 64.52 & \underline{52.69} & 58.14 \\
\ours+Ens & \underline{65.48} & 52.47 & \underline{62.79} \\
\ours+Ens+Aug & \textbf{65.80} & 51.54 & 62.59 \\
\bottomrule
\end{tabular}
}

\vspace{10pt}

{\scriptsize\textbf{(c) Inference compute}\par}
\vspace{1pt}

{\fontsize{5.7}{6.3}\selectfont
\setlength{\tabcolsep}{1.7pt}
\renewcommand{\arraystretch}{0.72}
\begin{tabular}{@{}lccc@{}}
\toprule
Method & GFLOPs & GMACs & Params \\
\midrule
CLIP & 8.7295 & 4.3623 & 87.46M \\
LoRA & 8.7885 & 4.3918 & 88.05M \\
\textbf{\ours} & 8.7342 & 4.3647 & 92.18M \\
\bottomrule
\end{tabular}
}

\end{minipage}

\end{table}

We evaluate \ours across nine image-classification datasets and two CLIP backbones. Our goal is not only to measure accuracy, but to understand when, why, and how SAE features can serve as a downstream steering space. We therefore organize the experiments around four questions: (i) can SAE sparse features improve zero-shot classification without labels, contrastive anchors, or test-time optimization? (ii) does the gain come from additive sparse steering rather than reconstruction alone? (iii) how much headroom remains if task-relevant sparse features can be selected? and (iv) can SAE reconstruction error predict when steering is unreliable under distribution shift?

We evaluate \ours on CIFAR-100~\citep{krizhevsky2009learning}, Tiny-ImageNet~\citep{le2015tiny}, CUB-200~\citep{wah2011caltech}, and six additional out-of-distribution datasets used for reliability analysis: Aircraft~\citep{aircraft}, Food101~\citep{food}, Flower102~\citep{flower}, Caltech101~\citep{caltech}, SUN~\citep{sun}, and EuroSAT~\citep{eurosat}. For the vision-language model, we use CLIP~\citep{radford2021learning} with ViT-B/32 and ViT-B/16 backbones. Unless otherwise specified, we train top-$k$ SAEs on unlabeled in-domain training-split activations and apply steering to the final-layer [CLS] token. Training details are provided in Appendix~\ref{sup:training}.

\begin{figure*}[t]
    \centering

    \begin{minipage}[c]{0.30\textwidth}
        \centering
        \includegraphics[width=\linewidth]{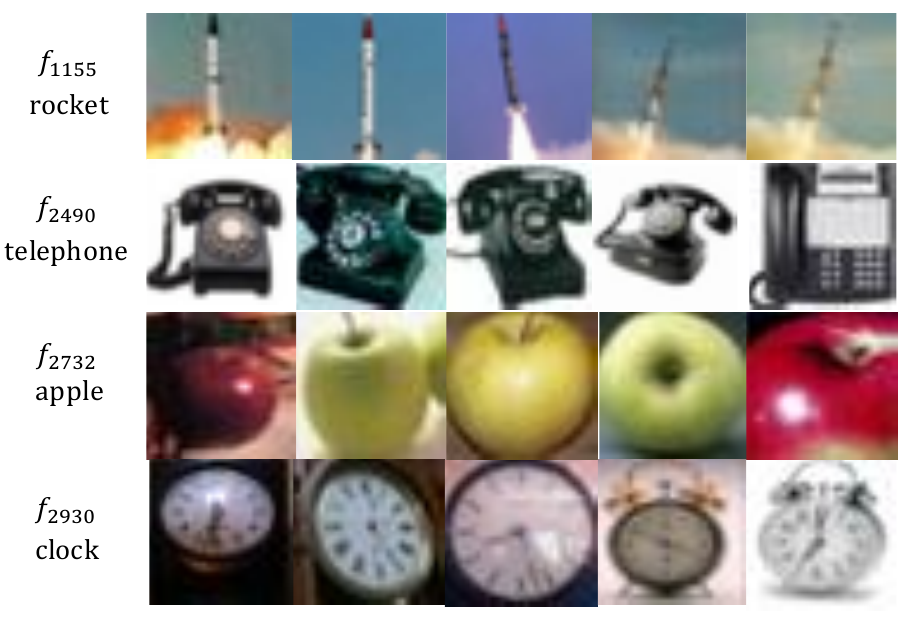}\\[-1mm]
        {\small\textbf{(a)} Interpretable SAE features.}
    \end{minipage}
    \hspace{0.03\textwidth}
    \begin{minipage}[c]{0.64\textwidth}
        \centering
        \includegraphics[width=\linewidth]{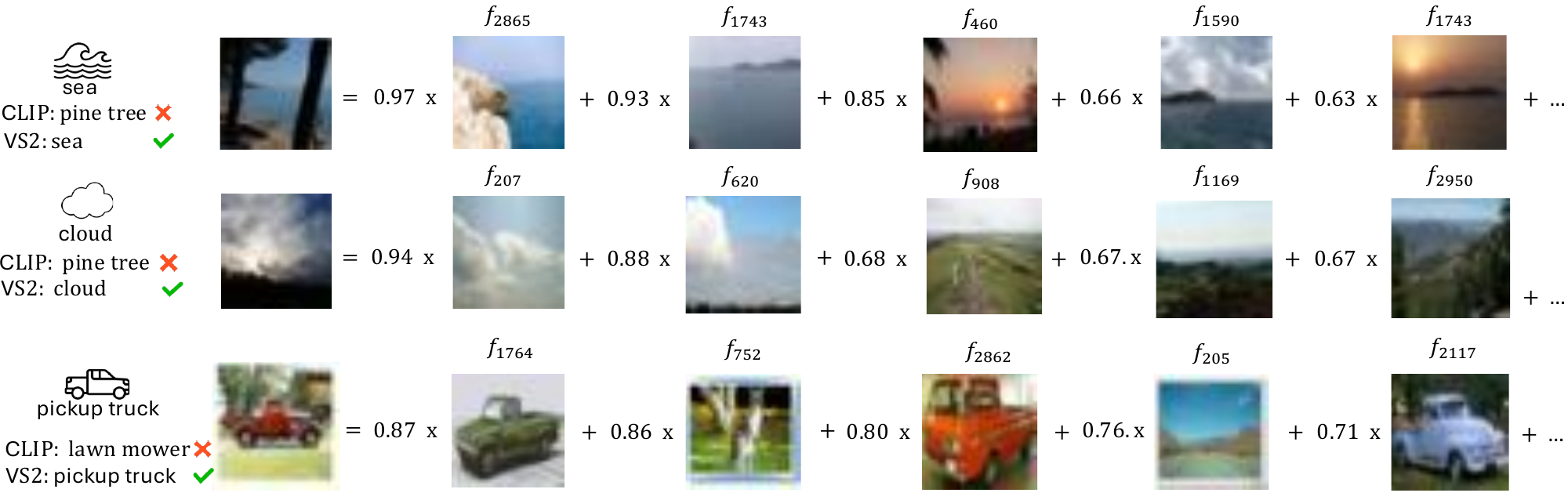}\\[-1mm]
        {\small\textbf{(b)} VS2-corrected examples using weighted interpretable features.}
    \end{minipage}

  \caption{
Qualitative evidence for SAE-based steering.
\textbf{(a)} Some SAE features group coherent visual concepts.
\textbf{(b)} In CIFAR-100 errors corrected by \ours, top active latents often
align with the recovered class, illustrating coarse semantic alignment rather
than perfect monosemanticity.
}
    \label{fig:vs2-qualitative-interpretability}
\end{figure*}

\subsection{\ours: Visual Sparse Steering}
\label{sec:benchmarking-ours}

We first evaluate whether SAE sparse features can improve zero-shot image classification without labels, contrastive anchors, or test-time optimization. This experiment also isolates the mechanism of \ours: if SAEs are used naively as reconstruction modules, performance may degrade due to reconstruction error; if sparse features are only amplified and reconstructed, the intervention is not consistently beneficial. \ours differs from these alternatives by using the SAE only to define an additive steering direction, while preserving the original CLIP embedding.

\textbf{Baselines.}
We compare \ours against baselines that isolate reconstruction and latent-amplification effects. We report: (1) the zero-shot CLIP model ($\text{CLIP}_{\text{ZS}}$); (2) $\text{SAE}_{\text{REC}}^{\text{F}}$, which replaces the final-layer [CLS] token with its SAE reconstruction; and (3) $\text{SAE}_{\text{REC}}^{\text{A}}$, which reconstructs [CLS] tokens from all layers. These baselines test whether SAE reconstruction alone improves classification. We additionally include $\text{SAE}_{\text{REC}}^{\text{F}+\gamma}$, which scales the top-$k$ latent activations by a fixed factor ($\gamma = 1.5$) before reconstructing the final-layer [CLS] token. This baseline tests whether latent amplification alone is sufficient, without the additive steering formulation of \ours. Appendix~\ref{appendix:pseudocode} provides pseudocode for these variants. We also report comparisons to Splice~\citep{bhalla2024interpreting} in Appendix~\ref{app:splice}, which defines latents using an external vocabulary rather than learning an unsupervised SAE dictionary. Unless stated otherwise, \ours applies steering to the final-layer [CLS] token; Appendix~\ref{sec:cls-layer-steering} analyzes steering across layers.

\begin{figure}[t]
  \centering
  \begin{subfigure}[t]{0.32\linewidth}
    \centering
    \includegraphics[width=\linewidth]{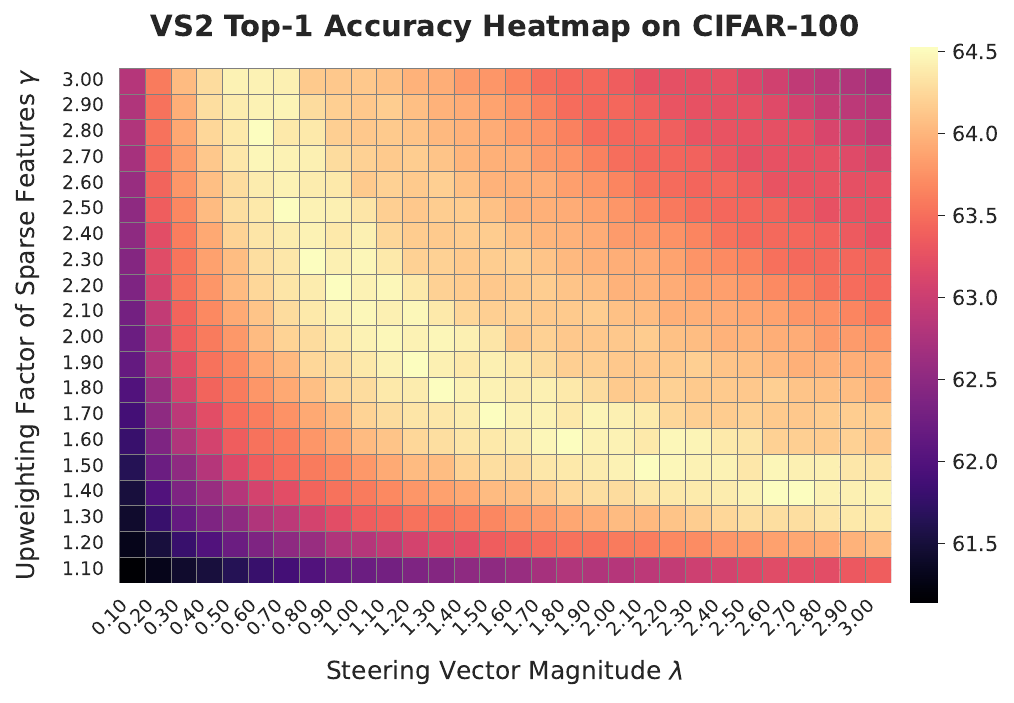}
    \label{fig:lambda-gamma-cifar}
  \end{subfigure}
  \hfill
  \begin{subfigure}[t]{0.32\linewidth}
    \centering
    \includegraphics[width=\linewidth]{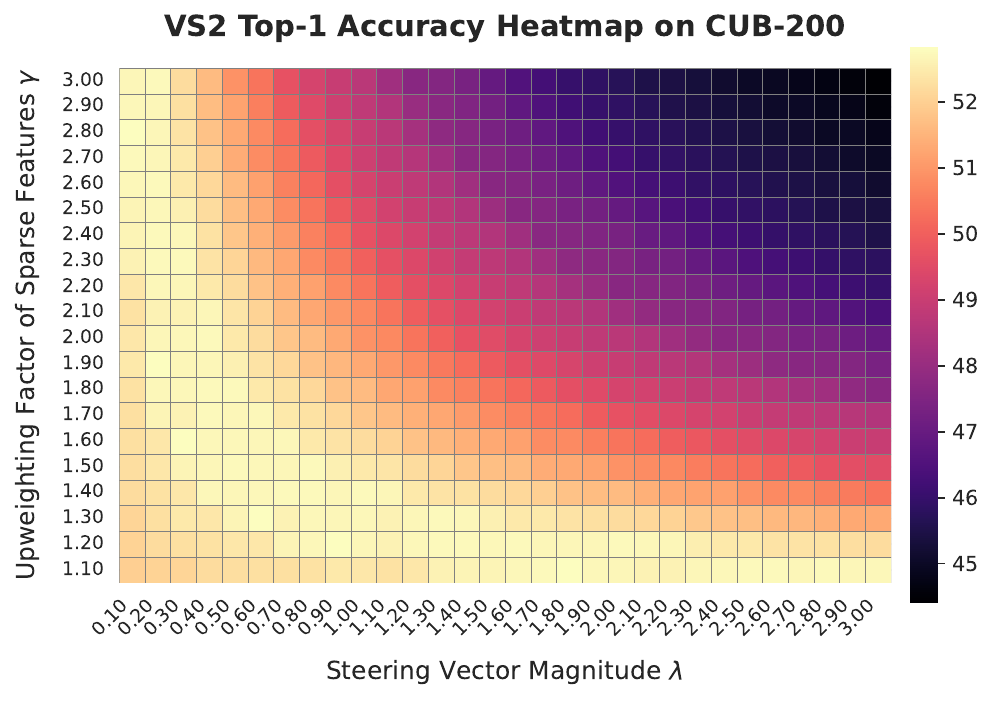}
    \label{fig:lambda-gamma-cub}
  \end{subfigure}
  \hfill
  \begin{subfigure}[t]{0.32\linewidth}
    \centering
    \includegraphics[width=\linewidth]{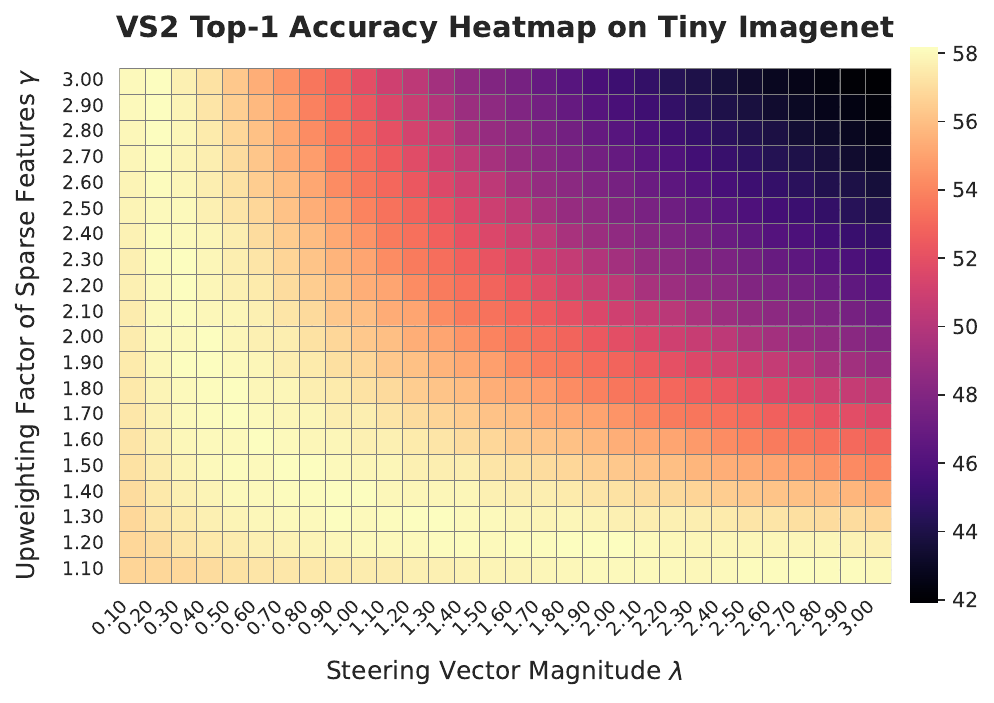}
    \label{fig:lambda-gamma-tiny}
  \end{subfigure}

  \caption{\textbf{Sensitivity of \ours{} to sparse amplification $\gamma$ and steering magnitude $\lambda$.}
  Across datasets, \ours exhibits a broad near-optimal region, typically when $\lambda\gamma \in [2,3]$.}
  \label{fig:lambda-gamma-heatmaps}
\end{figure}

\textbf{Results.}
Table~\ref{tab:acc_delta_inline} shows that SAE reconstruction alone usually hurts zero-shot performance, consistent with reconstruction error being directly injected into the CLIP embedding~\citep{engels2025decomposing}. Amplifying sparse features before reconstruction can help in some cases, such as CIFAR-100, where $\text{SAE}_{\text{REC}}^{\text{F}+\gamma}$ improves over zero-shot CLIP by 1.62\% and 2.85\% for ViT-B/32 and ViT-B/16, respectively. However, the same baseline underperforms on CUB-200 and Tiny-ImageNet, suggesting that amplified reconstruction is not a reliable intervention by itself.
In contrast, \ours consistently improves over zero-shot CLIP across all three datasets and both backbones. It yields gains of 3.45\% and 4.12\% on CIFAR-100, 0.93\% and 1.08\% on CUB-200, and 1.50\% and 1.84\% on Tiny-ImageNet for ViT-B/32 and ViT-B/16, respectively. These results show that the benefit comes not from replacing CLIP embeddings with SAE reconstructions, but from using sparse features to define an additive steering direction. Finally, Figure~\ref{fig:lambda-gamma-heatmaps} and Appendix~\ref{sec:sensitivity-lambda-gamma} analyze sensitivity to the steering hyperparameters $\gamma$ and $\lambda$. Across datasets, \ours exhibits a broad region of near-optimal performance, indicating that the method is not overly sensitive to precise hyperparameter choices.

 \textbf{Qualitative feature evidence.}
Although our main goal is downstream steering rather than interpretability alone, we inspect whether the sparse features used by \ours correspond to recognizable visual factors. Figure~\ref{fig:vs2-qualitative-interpretability} shows that individual SAE latents can group coherent visual concepts and that, in examples corrected by \ours, the active latents often align with coarse visual evidence for the recovered class. Additional qualitative examples and analyses are provided in Appendix~\ref{appendix:sae}.

\subsection{\oursplus: Upper-Bound Analysis via Selective Amplification}
\label{sec:benchmarking-oursplus}

The previous section shows that uniformly amplifying an input's active SAE features can improve zero-shot classification. However, this raises a natural question: are all active sparse features equally useful for the downstream task? Since the SAE is trained for reconstruction rather than classification, its active features may include both task-relevant evidence and nuisance factors such as background, texture, or pose. This creates a potential mismatch between \emph{reconstruction saliency} and \emph{task saliency}. To quantify the headroom available if task-relevant sparse features could be selected reliably, we introduce \oursplus as a controlled upper-bound analysis based on selective amplification.

Concretely, for each test image we retrieve the top-$50$ nearest neighbors from the dataset's unlabeled \emph{training split} using DINOv2~\citep{oquab2023dinov2}. We then form pseudo-positive and pseudo-negative groups using CLIP predictions: neighbors with the most frequent pseudo-label constitute positives, while the remaining neighbors are treated as negatives. We construct a contrastive steering direction in SAE feature space by subtracting the average negative sparse activations from the average positive sparse activations. This corresponds to the \emph{non-oracle} setting in Table~\ref{tab:acc_delta_inline}. For comparison, we also report an \emph{oracle} upper bound in which positives and negatives are defined using the ground-truth labels of the retrieved neighbors. We emphasize that \oursplus is not intended as a deployment-ready replacement for \ours. Instead, it is an analysis tool for measuring how much accuracy is left unrealized by uniform amplification. As baselines, we include a weighted Retrieval Augmented Generation (RAG) pipeline~\citep{lewis2020retrieval}, which combines the query embedding with a weighted average of retrieved CLIP embeddings (Appendix~\ref{sec:rag-ablation}), and a non-SAE contrastive steering baseline that computes the difference between the mean CLIP embeddings of the positive and negative groups. These comparisons distinguish the effect of retrieval alone, CLIP-space contrastive steering, and SAE-space selective amplification.

\textbf{Results.}
Table~\ref{tab:acc_delta_inline} reports \oursplus under \emph{oracle} and \emph{non-oracle} positive/negative sets. In the oracle setting, \oursplus achieves gains of up to 21.44\% on CIFAR-100, 7.08\% on CUB-200, and 20.47\% on Tiny-ImageNet. These large gains show that sparse feature selection has substantial headroom beyond uniform amplification, and support our claim that the main limitation of SAE-based downstream steering is not the absence of useful sparse features, but the difficulty of selecting the task-relevant subset.
Weighted RAG is generally weaker than contrastive steering on CIFAR-100 and Tiny-ImageNet while remaining competitive on CUB-200, suggesting that directional offsets are often more informative than proximity alone. \oursplus is competitive with CLIP-space steering, although it is not uniformly best, for example on Tiny-ImageNet with ViT-B/32. In the non-oracle setting, performance drops due to pseudo-label noise, but \oursplus still outperforms the CLIP steering-vector baseline in four of six configurations and avoids negative transfer on CUB-200. Overall, these results identify the \emph{reconstruction-vs-task saliency gap} as a central challenge for SAE-based downstream interventions and motivate future work on robust sparse-feature selection under weak supervision. Appendix~\ref{app:retrieved} ablates the number of retrieved neighbors.

\begin{table}[t]
\centering
\caption{\textbf{Top-5 class gains on CIFAR-100.} Top-1 accuracy with absolute gain over zero-shot CLIP. ``Mislabel'' denotes the most common zero-shot CLIP confusion corrected by each method.}
\label{tab:top5_class_gains}

\scriptsize
\setlength{\tabcolsep}{2.8pt}
\renewcommand{\arraystretch}{1.03}

\begin{minipage}[t]{0.49\linewidth}
\centering
\begin{tabular}{@{}lccL{1.35cm}L{2.25cm}@{}}
\toprule
\rowcolor{tableheader}
\multicolumn{5}{c}{\textbf{\ours}} \\
\rowcolor{tablesubheader}
Class & ZS & \ours & Mislabel & Visual confusion \\
\midrule
tractor & 55.0 & 80.0 \gain{(+25.0)} & lawn mower & wheeled farm machines \\
forest  & 35.0 & 58.0 \gain{(+23.0)} & pine tree & dense conifer canopy \\
man     & 53.0 & 74.0 \gain{(+21.0)} & boy & weak age cue at \(32{\times}32\) \\
bus     & 51.0 & 68.0 \gain{(+17.0)} & pickup & boxy vehicle silhouette \\
snake   & 61.0 & 76.0 \gain{(+15.0)} & worm & elongated legless body \\
\bottomrule
\end{tabular}
\end{minipage}
\hfill
\begin{minipage}[t]{0.49\linewidth}
\centering
\begin{tabular}{@{}lccL{1.35cm}L{2.25cm}@{}}
\toprule
\rowcolor{tableheader}
\multicolumn{5}{c}{\textbf{\oursplus}} \\
\rowcolor{tablesubheader}
Class & ZS & \oursplus & Mislabel & Visual confusion \\
\midrule
spider   & 53.0 & 91.0 \gain{(+38.0)} & bee & small dark silhouette \\
tiger    & 48.0 & 84.0 \gain{(+36.0)} & lion & stripes vs.\ mane unclear \\
flatfish & 49.0 & 85.0 \gain{(+36.0)} & whale & flat marine shape on blue \\
possum   & 30.0 & 65.0 \gain{(+35.0)} & hamster & small brown mammal \\
lizard   & 39.0 & 74.0 \gain{(+35.0)} & snake & legs often invisible \\
\bottomrule
\end{tabular}
\end{minipage}
\end{table}

% \subsection{Fine-Grained Per-Class Accuracy Analysis}
% \label{sec:per-class}
\textbf{Fine-grained class gains.}
Beyond aggregate accuracy, we analyze which classes benefit most from sparse steering. We compute per-class top-1 accuracy on \textsc{CIFAR-100} using a ViT-B/32 backbone and report the largest gains in Table~\ref{tab:top5_class_gains}. The largest improvements occur on visually confusable categories: \ours improves accuracy by up to 25\% absolute on classes such as \textit{tractor}, \textit{forest}, and \textit{man}, where the dominant zero-shot CLIP errors involve nearby visual or semantic alternatives such as \textit{lawn mower}, \textit{pine tree}, and \textit{boy}. With high-quality retrieved neighbors, \oursplus reaches gains of up to 38\% absolute on similarly confusable cases such as \textit{spider}, \textit{tiger}, and \textit{flatfish}.
This pattern suggests that sparse steering does not merely shift predictions uniformly across classes. Instead, it appears most helpful when the zero-shot decision is close to a visually related class and the intervention can amplify image-specific evidence that separates the correct class from its nearest confounder. This is consistent with the centroid-deviation view of \ours: steering away from shared population structure can emphasize distinctive visual attributes, while selective amplification in \oursplus can further improve performance when task-relevant sparse features are identified reliably. A full per-class breakdown is provided in Table~\ref{tab:vs2_vs2pp_gainloss} in Appendix~\ref{app:performance}.

\subsection{SAEs as a Reliability Diagnostic for Safe Visual Sparse Steering}
\label{sec:safe}

The previous experiments show that SAE features can improve classification when the SAE faithfully represents the target distribution. However, this condition is not guaranteed in deployment: a fixed SAE may be applied to samples whose activations differ from those seen during SAE training. In this case, the sparse features may no longer provide a faithful basis for steering, and the same intervention that improves accuracy in-domain can become harmful. We therefore ask whether the SAE's own reconstruction error can serve as a label-free diagnostic for deciding when to apply steering.

\textbf{Relaxation: a generalized SAE.}
Our main experiments use SAEs trained on unlabeled in-domain training-split activations for each dataset. This isolates the effect of sparse steering under favorable reconstruction conditions. To approximate a more realistic deployment setting, we train a single \emph{generalized} SAE on the union of the unlabeled training splits of CIFAR-100, CUB-200, and Tiny-ImageNet, and then apply this same SAE across datasets.
% Table~\ref{tab:main-diagnostic} reports top-1 accuracy together with Fraction of Variance Unexplained (FVU) for generalized steering. Larger FVU indicates poorer reconstruction, and FVU $>1$ means the SAE reconstructs worse than a mean predictor. Generalized steering improves over zero-shot CLIP on CIFAR-100 and Tiny-ImageNet, where reconstruction error is relatively low (FVU $\approx 0.21$ and $\approx 0.44$, respectively), but degrades on CUB-200, where reconstruction error is much higher (FVU $\approx 1.93$). This pattern supports the prediction from Proposition~\ref{prop:centroid_deviation}: when FVU is high, the residual term in the steering direction is large, making the intervention unreliable.
We measure reconstruction fidelity using Fraction of Variance Unexplained (FVU), where larger values indicate poorer reconstruction and FVU $>1$ means the SAE reconstructs worse than a mean predictor. With the generalized SAE, \ours improves on CIFAR-100 by +3.56\% and +4.26\% for ViT-B/32 and ViT-B/16, respectively, where FVU is low ($\approx 0.21$). It also improves on Tiny-ImageNet by +3.09\% and +2.75\%, where FVU is moderate ($\approx 0.44$). In contrast, \ours degrades on CUB-200 by -2.95\% and -5.84\%, where FVU is high ($\approx 1.93$). This supports the prediction from Proposition~\ref{prop:vs2-centroid}: high FVU corresponds to a larger residual magnitude in the steering direction and less reliable steering.

\begin{table*}[t]
\captionsetup{skip=2pt}
\centering
\caption{
\textbf{FVU-gated fallback under distribution shift.}
Each cell reports top-1 accuracy with coverage, i.e., the fraction of samples for which steering is applied, in parentheses. The gate disables steering when per-sample FVU exceeds a threshold calibrated without labels.
}
\label{tab:fvu_gate_all}
\scriptsize
\setlength{\tabcolsep}{3.5pt}
\renewcommand{\arraystretch}{0.95}
\resizebox{\linewidth}{!}{%
\begin{tabular}{l c c c c c c c c c}
\toprule
& \multicolumn{3}{c}{\textbf{In Distribution}} & \multicolumn{6}{c}{\textbf{Out of Distribution}} \\
\cmidrule(lr){2-4}\cmidrule(lr){5-10}
\textbf{Method} &
\textbf{CIFAR-100} & \textbf{CUB-200} & \textbf{Tiny-IN} &
\textbf{Aircraft} & \textbf{Food101} & \textbf{Flower102} & \textbf{Caltech101} & \textbf{SUN} & \textbf{EuroSAT} \\
\midrule
Baseline (CLIP$_{\text{ZS}}$) &
61.07 & 51.76 & 56.64 &
24.87 & 80.72 & 63.70 & 78.53 & 51.79 & 31.61 \\
VS2 (Target-domain SAE, ungated) &
64.52 & 52.69 & 58.14 &
27.69 & 81.20 & 64.81 & 80.95 & 54.80 & 34.65 \\
\midrule
VS2 (Generalized SAE, ungated) &
64.63 & 48.81 & 59.73 &
18.57 & 81.06 & 64.11 & 79.72 & 45.78 & 30.76 \\
VS2 + FVU gate ($q{=}0.90$) &
64.27 (99.93\%) & 51.76 (0.0\%) & 59.29 (85.0\%) &
24.87 (0.00\%) & 80.72 (0.00\%) & 63.70 (0.11\%) & 78.44 (2.30\%) & 51.80 (0.11\%) & 30.39 (54.20\%) \\
VS2 + FVU gate ($q{=}0.95$) &
64.62 (99.97\%) & 51.76 (0.04\%) & 58.44 (92.16\%) &
24.87 (0.00\%) & 80.71 (0.02\%) & 63.72 (1.32\%) & 78.52 (5.51\%) & 51.80 (0.49\%) & 31.39 (63.15\%) \\
VS2 + FVU gate ($q{=}0.99$) &
64.63 (100\%) & 51.79 (0.62\%) & 58.65 (98.63\%) &
24.87 (0.09\%) & 80.70 (0.89\%) & 64.34 (22.72\%) & 78.76 (23.82\%) & 51.92 (6.97\%) & 31.57 (85.52\%) \\
VS2 + FVU gate ($q{=}0.995$) &
64.63 (100\%) & 51.83 (1.54\%) & 58.64 (99.34\%) &
24.87 (0.21\%) & 80.69 (3.10\%) & 64.21 (40.62\%) & 78.78 (35.26\%) & 52.11 (15.28\%) & 31.65 (92.09\%) \\
\bottomrule
\end{tabular}
}
\end{table*}

\textbf{Out-of-distribution steering with FVU-gated fallback.}
We next evaluate whether FVU can prevent harmful steering under distribution shift. We apply the generalized SAE to six unseen datasets: Aircraft~\citep{aircraft}, Food101~\citep{food}, Flower102~\citep{flower}, Caltech101~\citep{caltech}, SUN~\citep{sun}, and EuroSAT~\citep{eurosat}. As an upper reference, Table~\ref{tab:fvu_gate_all} also reports VS2 with an SAE trained on each dataset's unlabeled training split. In this target-domain SAE setting, VS2 improves over zero-shot CLIP across all six datasets, suggesting that sparse steering remains effective when reconstruction is faithful.

By contrast, ungated generalized steering can substantially degrade performance on some shifted datasets, such as Aircraft and SUN, where the generalized SAE does not reconstruct the target activations reliably. We therefore use per-sample FVU as a label-free reliability diagnostic: if a test sample's FVU exceeds a threshold $\tau$, we skip steering and fall back to the original CLIP embedding; otherwise, we apply \ours.
To set $\tau$ without labels, we calibrate a single threshold on the unlabeled CIFAR-100+Tiny-ImageNet training union by taking the $q$-quantile of per-sample FVU scores, and then apply the same threshold across all datasets. For FVU-gated rows in Table~\ref{tab:fvu_gate_all}, we report both top-1 accuracy and \emph{coverage}, the fraction of samples for which steering is applied. FVU gating acts as a conservative safety mechanism: when generalized steering is harmful, coverage drops and performance returns toward the zero-shot baseline; when reconstruction is reliable, coverage remains high and most of the gains are retained. This gives \ours a model-internal reliability signal that standard steering-vector and test-time adaptation pipelines typically lack.

\section{Limitations and Future Work}
\label{sec:limitations}

\ours shows that SAE latents can serve not only as tools for post-hoc interpretation, but also as an actionable intervention space for downstream visual steering. However, the method has a genuine offline prerequisite: a top-$k$ SAE must be trained on activations representative of the target distribution. When the SAE generalizes poorly to a sample, Proposition~\ref{prop:vs2-centroid} shows that the steering residual is controlled by the per-sample FVU. Our FVU gate therefore trades coverage for safety by conservatively reverting to zero-shot CLIP when reconstruction is unreliable. This reduces harmful steering under distribution shift, but it also means that \ours may occasionally forgo potential gains. Future work could improve SAE fidelity at deployment through lightweight per-instance self-reconstruction or adaptive sparse coding before steering.

A second limitation is the mismatch between reconstruction and task saliency. SAEs are trained to reconstruct activations, so they surface features that are salient for reproducing the embedding rather than necessarily features that are optimal for classification. This reconstruction-vs-task saliency gap limits uniform sparse-feature amplification and motivates task-aligned sparse steering: selecting and amplifying the subset of SAE features that most directly supports the target decision. Appendix~\ref{app:prototype} takes an initial step in this direction with Prototype-Alignment Sparse Steering, inspired by prototype theory~\citep{rosch1973natural}, suggesting that weak or indirect task structure can better align sparse features with downstream needs.

Empirically, we evaluate \ours on CLIP ViT-B/32 and ViT-B/16 across nine datasets. ImageNet-scale benchmarks and non-CLIP vision-language models remain important directions for future evaluation. Finally, \ours inherits biases from the underlying vision-language model and amplifies features selected by the SAE. If these features are misaligned with human-relevant or safety-critical attributes, this may be difficult to detect in advance. Safety-critical use should therefore pair \ours with the FVU gate, per-class auditing, and task-specific validation.

\vspace{-2pt}
\section{Conclusion}

We introduced Visual Sparse Steering (\ours), a lightweight, label-free method for using Sparse Autoencoder (SAE) latents as an actionable intervention space in frozen vision-language models. \ours constructs per-instance steering directions from active sparse features, improving zero-shot image classification without labels, contrastive anchors, weight updates, or test-time optimization. Across 9 datasets, \ours consistently improves over zero-shot CLIP while adding less than $0.1\%$ inference compute.
Beyond empirical gains, we showed that \ours implements centroid-deviation steering, and offers an FVU-based reliability gate that disables steering when the SAE representation is unreliable, reducing harmful interventions under distribution shift. Finally, our selective-amplification study with \oursplus reveals substantial headroom when task-relevant sparse features can be identified, exposing the reconstruction-vs-task saliency gap as a central challenge for future SAE-based downstream methods. Overall, our results suggest that SAEs can move beyond interpretability and provide a practical, analyzable, and reliability-aware basis for label-free visual steering.

\bibliographystyle{plainnat}
\bibliography{icml2026bib}

\newpage
\appendix
\section*{Appendix}
\section{Decoding the Sparse Latent Space: Insights from SAEs}
\label{appendix:sae}
Our proposed methods achieve significant classification performance gains, primarily due to the contribution of Sparse Autoencoders (SAEs). We hypothesize that SAEs identify meaningful sparse features, which in turn guide the steering mechanisms. To validate this assumption, we conduct both quantitative and qualitative evaluations to assess the significance of the learned features.

\subsection{\textbf{Quantitative Evaluation of Feature Significance}} 
\label{appendix:quantitative}
To evaluate the role of sparse features, we manipulate the top-$k$ most active latent features extracted via Sparse Autoencoders. This experiment examines whether these features are critical for classification and how model predictions change under different modifications. We explore the following manipulation settings:
\begin{enumerate}
    \item \textbf{Zeroing-Out ($\gamma=0$)}: We set the top-$k$ most active features to zero before applying the steering vector. This removes their influence while preserving the remaining latent structure.
    \item \textbf{Negation ($\gamma=-1$)}: We invert the sign of the top-$k$ features before applying the steering vector, effectively pushing the representation in the opposite direction. This tests whether these dimensions encode class-discriminative information.
\end{enumerate}

\noindent
Table~\ref{tab:concept_manipulation} reports the zero-shot classification accuracy of $\text{SAE}_{\text{REC}}^{\text{F}+\gamma}$ using ViT-B/32 after applying these transformations to the Sparse Steering vector intervention. \textbf{We observe that negating or zeroing the top-$k$ most important features consistently degrades performance across all datasets}. This result confirms that the features learned by the Sparse Autoencoders are essential for classification.

\begin{table}[H]
\centering
\caption{\textbf{Effect of manipulating top-$k$ sparse codes.}
Zero-shot accuracy (\%) drops sharply when dominant sparse features are zeroed ($\gamma=0$) or negated ($\gamma=-1$), confirming their importance.}
\label{tab:concept_manipulation}
\small
\setlength{\tabcolsep}{5pt}
\begin{tabular}{l S S}
\toprule
\textbf{Modification} & {CIFAR-100} & {Tiny-IN} \\
\midrule
$\text{CLIP}_{\text{ZS}}$           & 61.07 &  56.64 \\
 $\text{SAE}_{\text{REC}}^{\text{F}+\gamma}$ & 62.69 &  39.49 \\
\addlinespace[2pt]
\textit{+ Zero-out ($\gamma=0$)} & 1.71  &  16.11 \\
\textit{+ Negate ($\gamma=-1$)}  & 0.06  &   0.82 \\
\bottomrule
\end{tabular}
\end{table}
\subsection{\textbf{Qualitative Evaluation of Features Significance}} 
\label{appendix:qualitative}
We qualitatively investigate the features learned in the sparse latent representations of the top-$k$ activations in the Sparse Autoencoder (SAE). Specifically, we assess the learned features by analyzing feature activations for each input and identifying the inputs that exhibit the highest activations for a given feature. Unlike mechanistic interpretability in the language domain, where an LLM can be used to assign semantic labels to a feature by summarizing its highly activated inputs, the vision domain lacks an equivalent automated labeling process. 

To avoid reliance on human qualitative evaluation, we leverage annotated datasets where each image is associated with predefined attributes. For the qualitative evaluation of feature significance, we use the CUB dataset, which provides rich concept annotations for each image, enabling a structured assessment of the learned representations. Specifically, we investigate whether we can identify specific latent features with the highest \textbf{concept coverage} among their top-k most activated images. Concept coverage refers to how consistently a specific interpretable concept (e.g., an identifiable object category, attribute, or semantic idea) appears across a set of highly activated examples for a given SAE dimension. The intuition is that if a specific concept frequently emerges among the top-activating images for a particular feature, that feature is strongly associated with  that concept. 
 
\begin{figure*}[hbtp]
  \centering
    \includegraphics[width=0.49\textwidth]{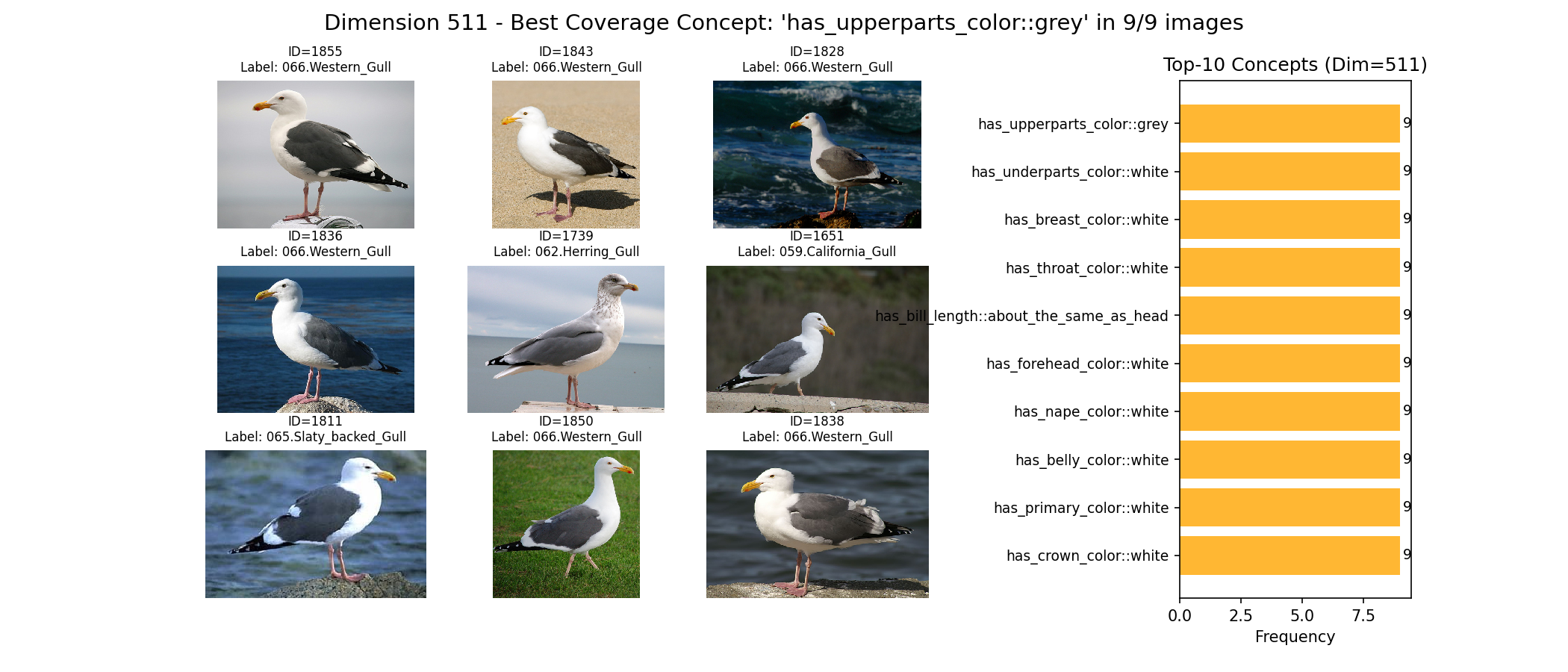}
    \includegraphics[width=0.49\textwidth]
    {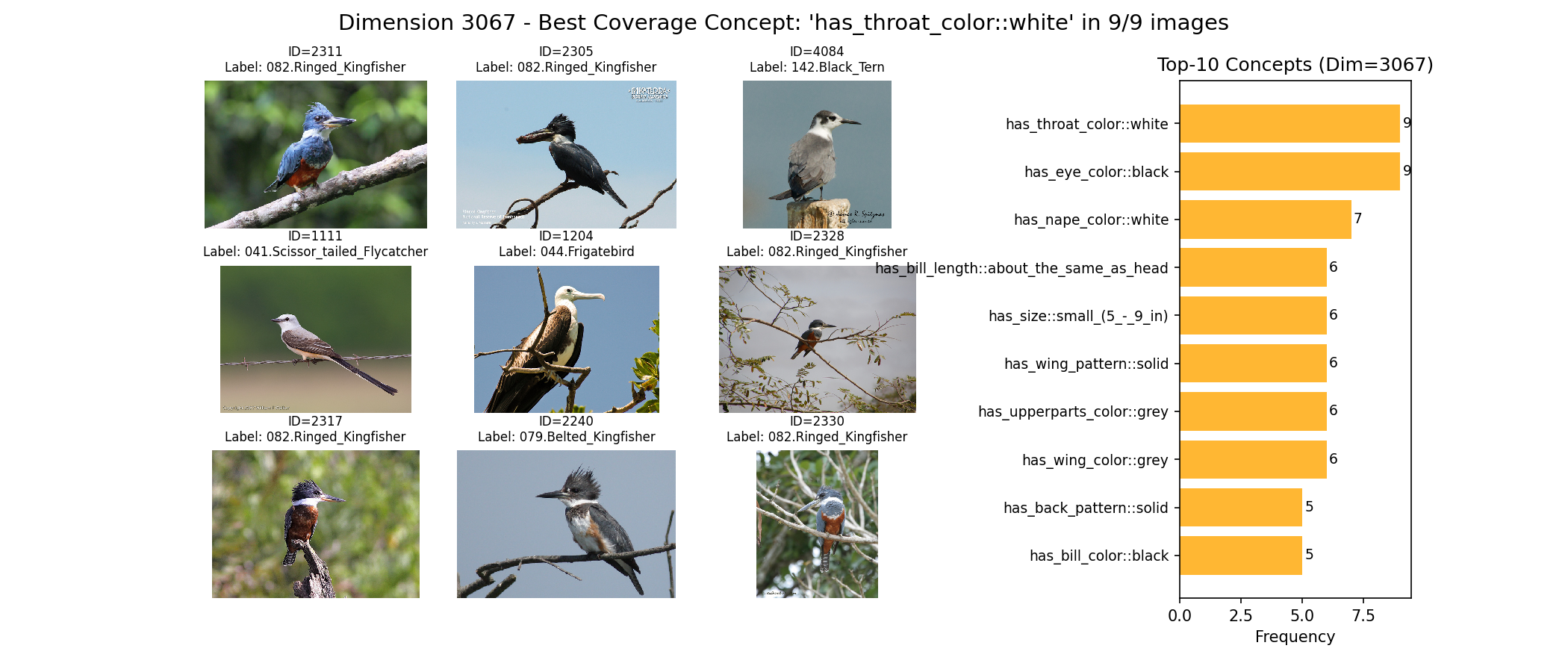}
  
\caption{Concept coverage analysis of learned sparse latent features in the Sparse Autoencoder (SAE). Each subfigure illustrates the top-activating images for two different SAE dimensions, highlighting the consistency of shared visual concepts among the highest-activated examples. The analysis demonstrates how sparse features capture meaningful semantic attributes, grouping semantically similar images either within or across classes. \textbf{Left}: Top-activating images for feature 511. The images predominantly belong to different classes of Gulls (e.g., Western, California, Herring, and Slaty-Backed Gulls), yet they share consistent visual characteristics such as a gray upper part and a white underpart. This suggests that feature 511 captures a semantically meaningful concept spanning multiple related categories. \textbf{Right}: Top-activating images for feature 3067. The images share distinct visual attributes, including a white-colored throat and black eyes. However, unlike feature 511, these images belong to more diverse categories, indicating that this latent dimension captures a broader concept that generalizes across different classes.}
\label{fig:two_images_side_by_side}
\end{figure*}

 For feature 511 as shown in Figure~\ref{fig:two_images_side_by_side} (left), the activated images exhibit consistent semantic characteristics, including a gray upper part and a white underpart. Notably, this feature predominantly activates for images from similar but different classes, specifically different types of Gulls, such as Western, California, Herring, and Slaty-Backed Gulls. We observe that the top-activating images for these classes share all concepts in common which is not the case with dimension 3067. For feature 3067, as shown in  Figure~\ref{fig:two_images_side_by_side} (right) we observe that the top-activating images share common visual attributes, such as a white-colored throat and black eyes.  Through human qualitative evaluation, we find that \textbf{features in the sparse latent space capture meaningful visual concepts, grouping semantically similar images together, either from the same class or across different classes, as long as they share underlying conceptual similarities}.

\begin{table}[h]
\centering
\caption{Top-10 most activated classes for SAE Feature 511 (left) and Feature 3067 (right). Feature 511 strongly aligns with gull-like classes, while Feature 3067 captures a cross-class head/throat attribute.}
\label{tab:feature_511_3067_split}

% \resizebox{\columnwidth}{!}{%
\begin{adjustbox}{max width=\columnwidth}
\begin{tabular}{l c l c}
\toprule
\textbf{Class (511)} & \textbf{Activation} & \textbf{Class (3067)} & \textbf{Activation} \\
\midrule
066.Western\_Gull & 96.7\% (29/30) & 082.Ringed\_Kingfisher & 60.0\% (18/30) \\
060.Glaucous\_winged\_Gull & 96.6\% (28/29) & 083.White\_breasted\_Kingfisher & 40.0\% (12/30) \\
061.Heermann\_Gull & 93.3\% (28/30) & 079.Belted\_Kingfisher & 30.0\% (9/30) \\
059.California\_Gull & 90.0\% (27/30) & 041.Scissor\_tailed\_Flycatcher & 23.3\% (7/30) \\
062.Herring\_Gull & 83.3\% (25/30) & 080.Green\_Kingfisher & 16.7\% (5/30) \\
063.Ivory\_Gull & 76.7\% (23/30) & 002.Laysan\_Albatross & 13.3\% (4/30) \\
087.Mallard & 76.7\% (23/30) & 025.Pelagic\_Cormorant & 13.3\% (4/30) \\
147.Least\_Tern & 76.7\% (23/30) & 025.Pelagic\_Cormorant & 13.3\% (4/30) \\
064.Ring\_billed\_Gull & 70.0\% (21/30) & 044.Frigatebird & 10.0\% (3/30) \\
084.Red\_legged\_Kittiwake & 73.9\% (17/23) & 049.Boat\_tailed\_Grackle & 10.0\% (3/30) \\
\bottomrule
\end{tabular}%
% }
\end{adjustbox}
\end{table}

\paragraph{Class-Conditional Feature Activation.} To provide additional evidence that SAE dimensions capture discriminative visual concepts, we report the class-conditional activation frequencies of two example features: 511 and 3067. Specifically, Table~\ref{tab:feature_511_3067_split} shows the percentage of samples in the top-10 activating classes where each feature is present.

What we observe is that feature 511 fires very often in gull-like classes. It also fires Larus-type seabirds, i.e., terns, kittiwakes. We observe that it also fires moderately on similar seabirds (Horned Puffin: 53.3\%, White Pelican: 40\%, Brown Pelican: 16.7\%, Red-breasted Merganser: 10\%). Additionally, we observe that Feature 511 activates near 0\% on unrelated classes such as buntings, warblers, and hummingbirds. In contrast, Feature 3067 captures a cross-category visual attribute rather than a taxonomic grouping. Specifically, it consistently activates on species that share a distinctive head-and-throat pattern (e.g., white throat with darker eye region), even when those species are taxonomically unrelated. This indicates that the SAE possibly learns attribute-level discriminative concepts, complementing the class-specific features exemplified by Feature 511.

\paragraph{Discussion.} We note that our goal is not to claim human interpretability of SAE features. As shown in recent mechanistic interpretability work \citep{chanin2409absorption}, human-aligned features are often split across multiple SAE components (feature absorption), a known behavior in sparse dictionary learning. Nonetheless, our contribution lies in demonstrating that SAEs can go beyond interpretability, serving as effective mechanisms for performance improvement via sparse steering. While some features are interpretable, our main result is that sparse features enable controllable improvements in downstream classification tasks.

\section{Sensitivity to Sparse Amplification ($\gamma$) and Steering Magnitude ($\lambda$)}
\label{sec:sensitivity-lambda-gamma}

We analyze how the two key hyperparameters in \ours (i) the sparse-feature amplification $\gamma$, and (ii) the steering vector scale $\lambda$ affect downstream accuracy. In the absence of contrastive supervision, these parameters govern how strongly we amplify sparse activations and how far the embedding is shifted in feature space. We sweep both values across a grid on three datasets: CIFAR-100, CUB-200, and Tiny-ImageNet using ViT-B/32 backbone.

\begin{figure*}[t]
  \centering
  \begin{subfigure}[t]{0.32\linewidth}
    \centering
    \includegraphics[width=\linewidth]{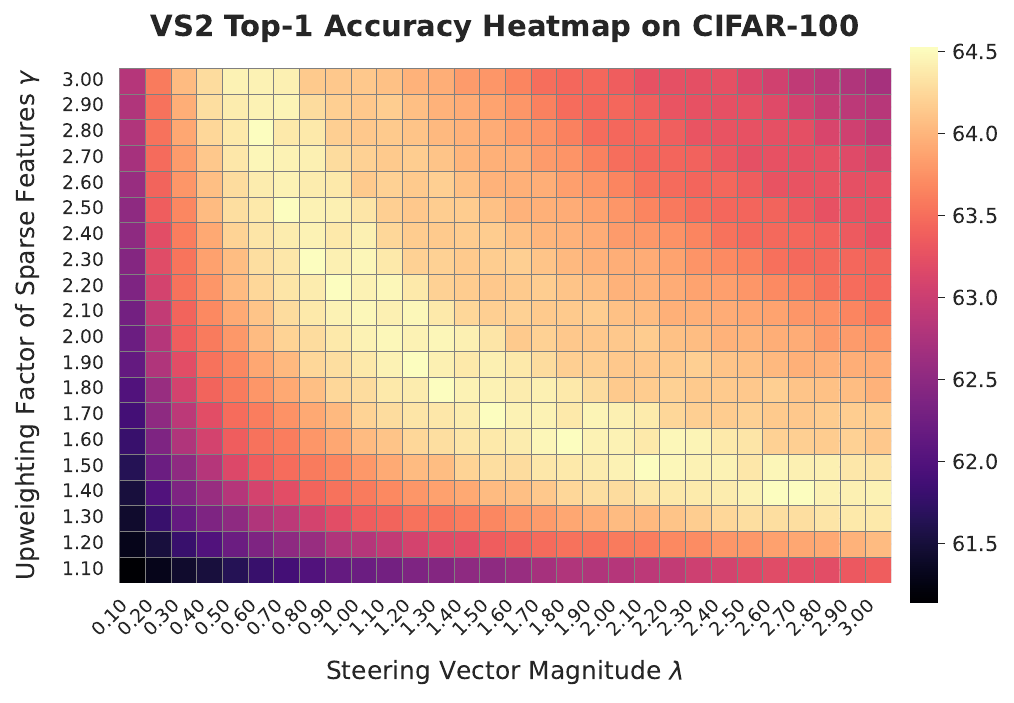}
    \caption{CIFAR-100}
    \label{fig:lambda-gamma-cifar}
  \end{subfigure}
  \hfill
  \begin{subfigure}[t]{0.32\linewidth}
    \centering
    \includegraphics[width=\linewidth]{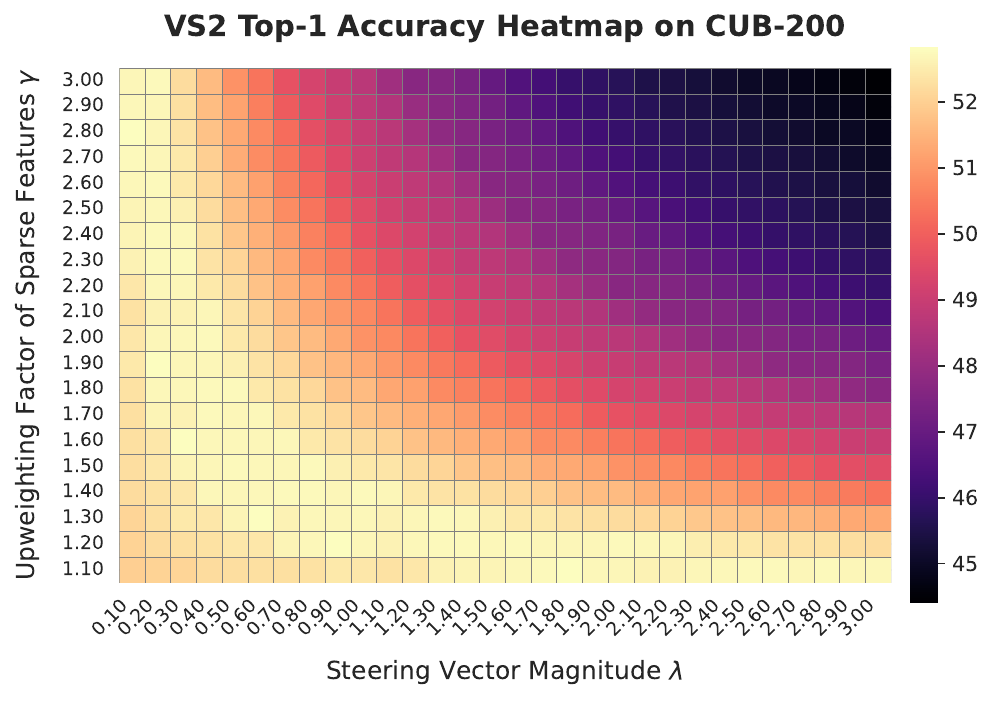}
    \caption{CUB-200}
    \label{fig:lambda-gamma-cub}
  \end{subfigure}
  \hfill
  \begin{subfigure}[t]{0.32\linewidth}
    \centering
    \includegraphics[width=\linewidth]{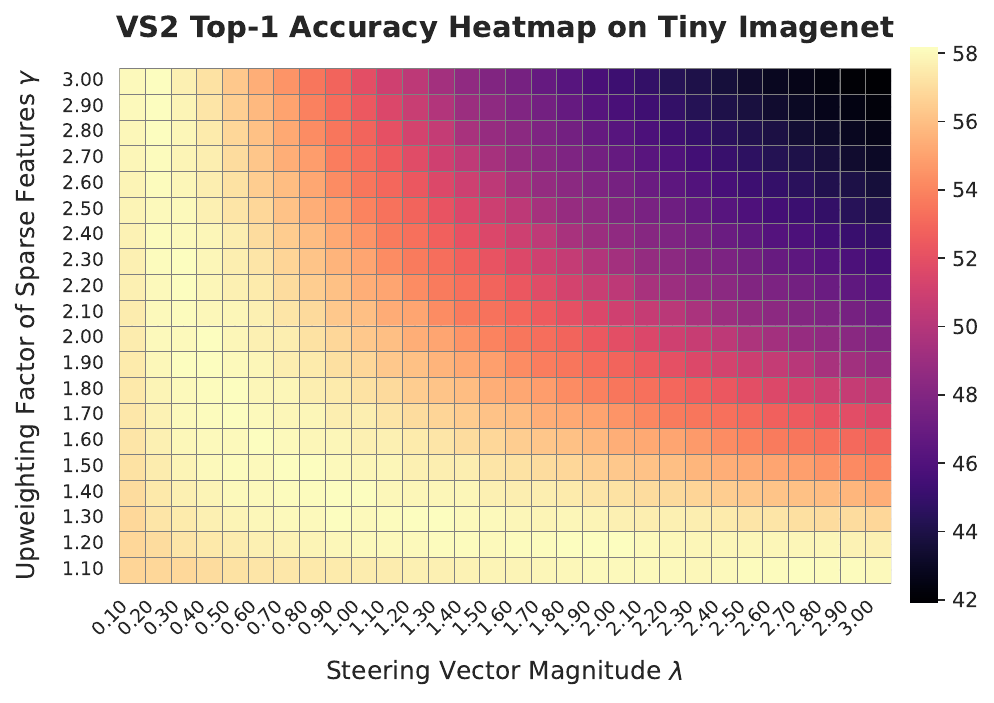}
    \caption{Tiny-ImageNet}
    \label{fig:lambda-gamma-tiny}
  \end{subfigure}
  \caption{\textbf{Sensitivity of \ours{} to sparse amplification $\gamma$ and steering magnitude $\lambda$}.  
  All three datasets show a range of near-optimal combinations (warm colours), typically when $\lambda \cdot \gamma \in [2, 3]$. Accuracy degrades if either parameter becomes too large.}
  \label{fig:lambda-gamma-heatmaps2}
\end{figure*}

Figure~\ref{fig:lambda-gamma-heatmaps2} shows that all tested combinations of $(\lambda, \gamma)$ outperform the zero-shot baseline, though to varying degrees. Each dataset exhibits a diagonal band of near-optimal settings where $\lambda \cdot \gamma \in [2, 3]$ tends to yield peak accuracy. For example, CIFAR-100 peaks at $\lambda^*=2.1$ and $\gamma^*=1.5$. Beyond this band, increasing $\lambda$ or $\gamma$ causes performance to degrade likely due to over-amplification of sparse features and/or embedding distortion. The consistent contour patterns across datasets suggest that \ours is robust to moderate variation in its hyperparameters and that good settings generalize well across domains.

\section{Implementation, Training, and Compute Details}
\label{sup:training}
We follow CLIP~\citep{radford2021learning} with a ViT-B/32 and ViT-B/16 vision backbones, 
intercepting the output of each encoder layer for the $CLS$ token. 
Specifically, we train top-$k$ SAEs on the $CLS$ embeddings for each chosen layer. 
We use $k=64$ and $k=256$ as the maximum active features within the latent space for ViT-B/32 and ViT-B/16 respectively, 
and we set a ``dead feature'' threshold of 100 i.e., any feature seldom activated is pruned. 
We also use an expansion factor of 4 relative to the input embedding dimension, 
resulting in 3{,}072 latent units. 
Training largely follows~\citet{gao2024scaling} and uses ~\citet{sparsify2024}, 
with a linear learning-rate schedule and warmup from $5\times10^{-4}$.

We monitor reconstruction quality with the \textbf{fraction of variance unexplained}
(\textbf{FVU}; lower is better), defined as
\begin{equation}
\text{FVU}
\;=\;
\frac{\lVert \mathbf{X} - \hat{\mathbf{X}} \rVert_{F}^{2}}
     {\lVert \mathbf{X} - \bar{\mathbf{X}} \rVert_{F}^{2}},
\label{eq:fvu}
\end{equation}
where $\mathbf{X} \in \mathbb{R}^{B \times d}$ is a batch of CLS embeddings,
$\hat{\mathbf{X}}$ denotes their SAE reconstructions, and
$\bar{\mathbf{X}}$ is the batch mean (so the denominator equals the total variance).
FVU is the complement of the coefficient of determination ($1 - R^{2}$); an
\mbox{FVU} of $0$ indicates perfect reconstruction, while an \mbox{FVU} of $1$
corresponds to predicting only the mean. In Table~\ref{tab:fvu_all}, we report FVU results for ViT-B/32 with $k=64$ and ViT-B/16 with $k=256$ across all three datasets.

\begin{table}[t]
\centering
\caption{\textbf{Fraction of variance unexplained (FVU; lower is better).}
Each SAE has 4× expansion; sparsity $k=64$ for ViT-B/32 and $k=256$ for
ViT-B/16.}
\label{tab:fvu_all}
\small
\setlength{\tabcolsep}{8pt}
\begin{tabular}{lSS}
\toprule
\multicolumn{1}{c}{Dataset} & 
\multicolumn{1}{c}{ViT-B/32\,(k=64)} &
\multicolumn{1}{c}{ViT-B/16\,(k=256)} \\
\midrule
CIFAR-100      & 0.2812 & 0.1166 \\
CUB-200        & 0.2487           & 0.1653 \\
Tiny-ImageNet  & 0.5060           & 0.3018 \\
\bottomrule
\end{tabular}
\end{table}

\paragraph{Compute resources.}
Experiments were run on an internal GPU cluster node with 8 NVIDIA Quadro RTX 8000 GPUs, each with 46{,}080 MiB of memory, using CUDA 12.6. SAE training and VS2 evaluation were run on a single GPU per experiment unless otherwise specified. VS2 inference adds only one SAE encode/decode pass on top of CLIP inference; Table~\ref{tab:acc_delta_inline} and Appendix~\ref{app:compute} report the inference-time and training-time compute, respectively.

\paragraph{Existing assets and licenses.}
We use publicly available pretrained models and image-classification datasets for standard research evaluation. We cite the original sources for all assets used in the paper, including CLIP, DINOv2, CIFAR-100, CUB-200, Tiny-ImageNet, Aircraft, Food101, Flower102, Caltech101, SUN, and EuroSAT. We use these assets only for their intended research and benchmark evaluation purposes and do not redistribute dataset contents or pretrained model weights.

\section{Sensitivity in Retrieval-Augmented Generation (RAG)}
\label{sec:rag-ablation}

\begin{figure*}[t]
  \centering
  \begin{subfigure}[t]{0.48\textwidth}
    \includegraphics[width=\linewidth]{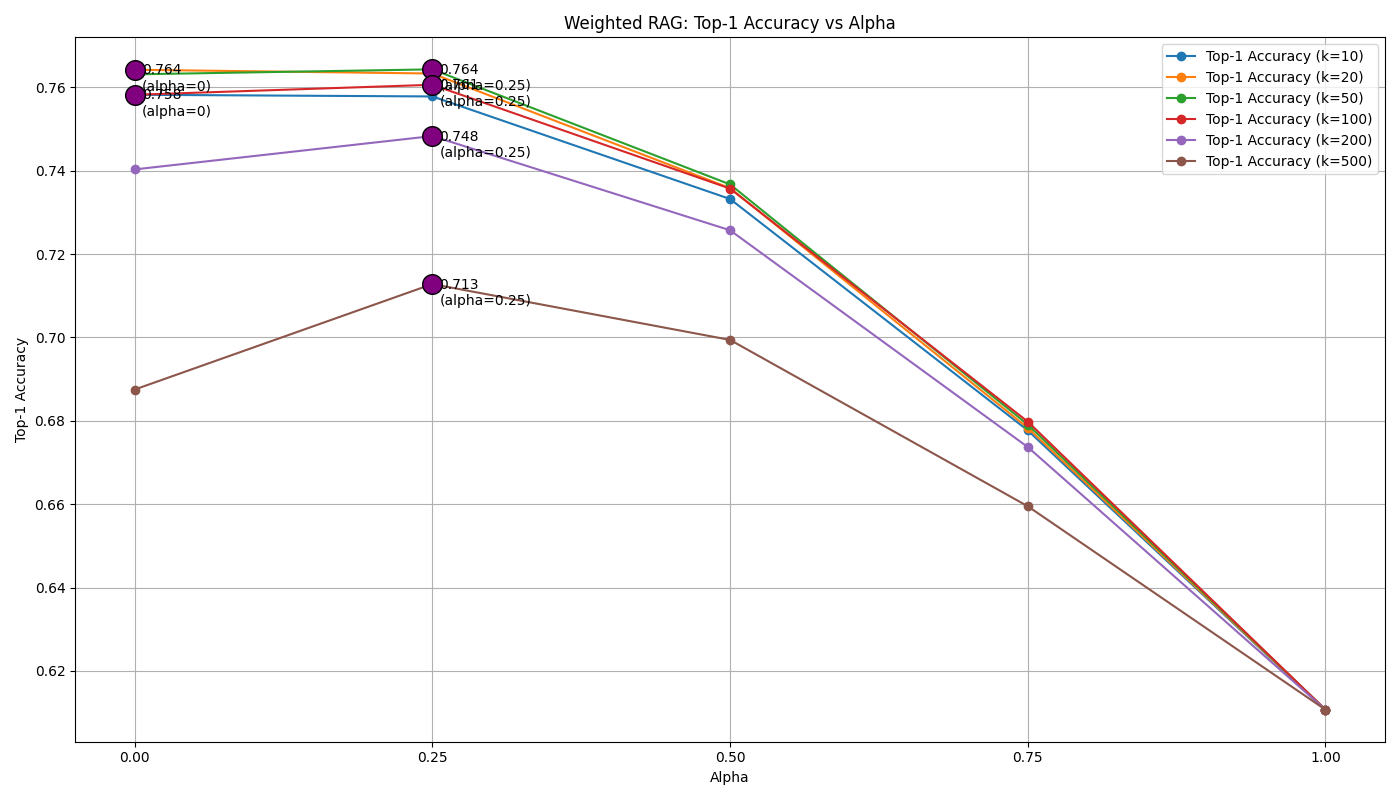}
    \caption{CIFAR-100}
  \end{subfigure}
  \hfill
  \begin{subfigure}[t]{0.48\textwidth}
    \includegraphics[width=\linewidth]{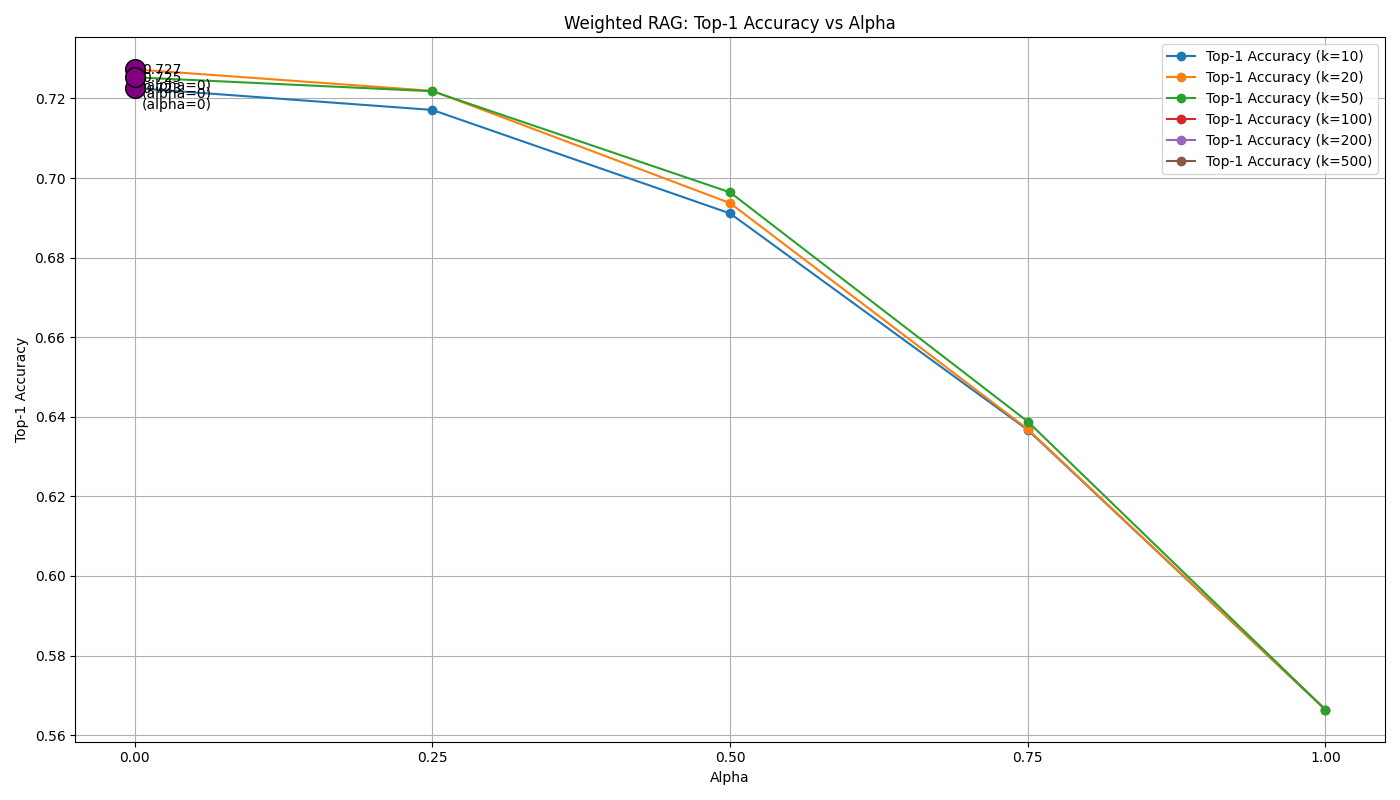}
    \caption{Tiny-ImageNet}
  \end{subfigure}
  \caption{\textbf{RAG sensitivity to $\alpha$ and top-$k$.}  
  Accuracy varies with the weight $\alpha$ on the original query and the number $k$ of retrieved images.  
  Larger $k$ often introduces noise; smaller $\alpha$ performs better on cluttered datasets.}
  \label{fig:rag-hyper}
\end{figure*}

When an external image corpus is available, \ours{} can be extended using a Retrieval-Augmented Generation (RAG) pipeline.  
We use DINOv2~\cite{oquab2023dinov2} to retrieve top-$k$ images most similar to the input query and compute an enhanced embedding by combining the query with the retrieved set:

\begin{equation*}
    \mathbf{E} = \alpha \mathbf{q} + (1 - \alpha) \sum_{j=1}^{k} w_j \mathbf{r}_j,
\end{equation*}

where $\mathbf{q}$ is the query embedding, $\mathbf{r}_j$ the $j$-th retrieved embedding, and $w_j$ the normalized similarity weight:
\begin{equation*}
    w_j = \frac{s_j}{\sum_{i=1}^{k} s_i}, \quad \text{where } s_j = \text{sim}(\mathbf{q}, \mathbf{r}_j).
\end{equation*}
The parameter $\alpha \in [0,1]$ controls the trade-off between using the original query and the retrieved set. We sweep over values of $\alpha$ and $k$ to assess their impact on zero-shot classification performance.  
Figure~\ref{fig:rag-hyper} shows results for CIFAR-100, and Tiny-ImageNet. CIFAR-100 and Tiny-ImageNet display a similar trend: larger $\alpha$ (i.e., more reliance on the query) typically degrades performance.  
On Tiny-ImageNet, setting $\alpha=0$, completely ignoring the input query and relying purely on retrieved features, yields the best result.  
%This suggests that the retrieval set carries strong distributional signal in cluttered datasets.  
Across all datasets, large $k$ eventually hurts performance, confirming that RAG benefits from focused rather than broad context. The trade-off parameter $\alpha$ and the retrieval depth $k$ play dataset-dependent roles in RAG-enhanced pipelines.  
For fine-grained domains, fewer, high-confidence neighbors and a low $\alpha$ might work best; for noisy domains, more reliance on the retrieval images might be more beneficial.

\section{Steering Across Layers of the CLS Token}
\label{sec:cls-layer-steering}

Our default method reconstructs and applies steering to the CLS token embedding at the final layer of the vision encoder. To study depth effects, we evaluate steering the CLS token across multiple final layers of the transformer. We use ViT-B/16 on CIFAR-100 with expansion factor~4 and 128 sparse activations. Table~\ref{tab:layer-steering} reports top-1 accuracy when steering the last 1, 2, or 3 layers. Steering only the final layer yields the best performance. Adding earlier layers causes a progressive drop in accuracy, falling below the zero-shot baseline when steering the last three layers.

\begin{table}[h]
\centering
\caption{\textbf{Effect of steering depth on CLS.} Top-1 accuracy (\%) on CIFAR-100 (ViT-B/16) as the number of steered final layers increases.}
\label{tab:layer-steering}
\small
\begin{tabular}{l c}
\toprule
\textbf{Steered Layers} & \textbf{Accuracy (\%)} \\
\midrule
12              & 68.08 \\
11 + 12         & 65.72 \\
10 + 11 + 12    & 59.36 \\
\bottomrule
\end{tabular}
\end{table}

Applying steering at multiple layers likely introduces compounding perturbations that propagate forward, making later representations harder to align with the classifier. 
CLS steering is most effective at the final layer. Future work could explore coordinated multi-layer steering to avoid error accumulation.

\section{Ablation Study of Expansion factor and top-$k$}
In Table \ref{tab:sae_width_k}, we present the downstream task accuracy of CLIP ViT-B/32 using various values of expansion factor and $k$. Across a 4\,$\times$ range in width and an 8\,$\times$ range in sparsity,
top-1 accuracy fluctuates by less than one percentage point 
\emph{evidence that VS2’s performance is largely insensitive to the precise SAE
capacity–sparsity trade-off.} Additionally, in Table \ref{tab:sae-sim-no-clip}, we present the average cosine similarities of different steering vectors coming from various configurations of SAEs in terms of expansion factor and top-$k$.
\label{app:expansion_k}

\begin{table}[h]
\centering
\caption{\textbf{VS2 accuracy as a function of SAE width (expansion factor) and sparsity (top-$k$).}
Numbers are top-1 / top-5 (\%). The best result is boldfaced; every other
configuration is within  of the optimum, highlighting the method’s robustness to architectural choices.}
\label{tab:sae_width_k}
\small
\setlength{\tabcolsep}{8pt}
\begin{tabular}{@{}l cc@{}}
\toprule
\multicolumn{1}{c}{\textbf{SAE configuration}}
      & \textbf{Top-1 $\uparrow$}
      & \textbf{Top-5 $\uparrow$} \\ 
\midrule
4\,$\times$, $k$=128                & \textbf{64.61} & \textbf{87.95} \\
8\,$\times$, $k$=128                & 64.56 & 87.76 \\
4\,$\times$, $k$=64                 & 64.54 & 87.79 \\
16\,$\times$, $k$=64                & 64.54 & 87.78 \\
8\,$\times$, $k$=64       & 64.52 & 87.96 \\
10\,$\times$, $k$=128               & 64.43 & 87.81 \\
16\,$\times$, $k$=512               & 64.42 & 87.71 \\
8\,$\times$, $k$=512                & 64.40 & 87.91 \\
4\,$\times$, $k$=256                & 64.34 & 87.62 \\
8\,$\times$, $k$=256                & 64.29 & 87.80 \\
16\,$\times$, $k$=256               & 64.28 & 87.75 \\
4\,$\times$, $k$=512                & 64.12 & 87.87 \\
16\,$\times$, $k$=128               & 64.10 & 87.79 \\
\bottomrule
\end{tabular}
\end{table}

\section{Visual Sparse Steering Pseudocode}
\label{appendix:pseudocode}
For reproducibility purposes, in Algorithm \ref{alg:sae_steering}, we provide the pseudocode for the baselines and \ours used in the analysis of \ref{sec:benchmarking-ours}. 
\label{pseudocode}
\begin{algorithm}[H]
\caption{\textsc{SAE\_Steering} -- SAE-based CLS token modification}
\label{alg:sae_steering}
\begin{algorithmic}[1]
\Function{SAE\_Steering}{$\mathbf{h}, \mathrm{SAE}, k, \gamma, \lambda_{\Delta z}, \text{mode}$}
    \Comment{$\mathbf{h}$: CLS token; SAE: sparse autoencoder; $k$: sparsity level}

    \State $\mathbf{a}, \mathbf{idx} \gets \mathrm{SAE.select\_topk}(\mathrm{SAE.pre\_acts}(\mathbf{h}), k)$
    \State $\mathbf{z}_{\text{base}} \gets \mathrm{SAE.decode}(\mathbf{a}, \mathbf{idx})$
    \State $\mathbf{z}_{\text{boost}} \gets \mathrm{SAE.decode}(\gamma \cdot \mathbf{a}, \mathbf{idx})$

    \If{$\text{mode} = \texttt{"reconstruction"}$}
        \State \Return $\mathbf{z}_{\text{base}}$ \Comment{Variant: $\text{CLIP}_{\text{REC}}^{\text{F}}$}
    \ElsIf{$\text{mode} = \texttt{"amplified"}$}
        \State \Return $\mathbf{z}_{\text{boost}}$ \Comment{Variant: $\text{CLIP}_{\text{REC}}^{\text{F}+\gamma}$}
    \ElsIf{$\text{mode} = \texttt{"steering"}$}
        \State \Return $\mathbf{h} + \lambda_{\Delta z} \cdot (\mathbf{z}_{\text{boost}} - \mathbf{z}_{\text{base}})$ \Comment{\ours}
    \EndIf
\EndFunction
\end{algorithmic}
\end{algorithm}

\section{Scaling to ViT-L/14}
\label{sec:appendix-vitl14}

To test whether VS2 transfers beyond the ViT-B/32 and ViT-B/16
backbones used in the main experiments, we apply the same steering
pipeline to the larger CLIP ViT-L/14 model
(\texttt{openai/clip-vit-large-patch14}; vision tower depth 24,
hidden size 1024). For each dataset, we train a dataset-specific
top-$k$ SAE on layer~23, corresponding to the final vision block, with
expansion factor $E{=}10$ and $k{=}128$ active latents. Steering is
applied only at the last layer (\texttt{last\_n}{=}1).

Table~\ref{tab:vitl14-summary} reports zero-shot Top-1 accuracy on the
three datasets used in the main paper. VS2 improves Top-1 accuracy on
all three datasets, suggesting that the proposed steering procedure
extends to larger CLIP vision backbones. The gains are smaller than
those observed for ViT-B/16, which is expected because ViT-L/14 starts
from a stronger zero-shot baseline; nevertheless, the consistent
positive deltas indicate that SAE-based steering is not specific to
the ViT-B family.

\begin{table}[t]
  \centering
  \caption{Scaling VS2 to CLIP ViT-L/14. We report zero-shot Top-1
  accuracy on the three main datasets. Each row uses a dataset-specific
  SAE trained on layer~23 with expansion factor $E{=}10$ and $k{=}128$,
  with steering applied only at the last layer
  (\texttt{last\_n}{=}1).}
  \label{tab:vitl14-summary}
  \small
  \begin{tabular}{llrrr}
    \toprule
    Dataset & Eval split ($n$) & CLIP ZS & VS2 & $\Delta$ \\
    \midrule
    CUB-200-2011  & test ($5\,794$)  & 62.50 & \textbf{64.26} & \textbf{+1.76} \\
    CIFAR-100     & test ($10\,000$) & 72.48 & \textbf{74.12} & \textbf{+1.64} \\
    Tiny-ImageNet & val  ($10\,000$) & 72.58 & \textbf{73.52} & \textbf{+0.94} \\
    \bottomrule
  \end{tabular}
\end{table}

\section{Comparison with other Baseline Methods}
\label{app:splice}
We compare \ours against SpLiCE~\citep{bhalla2024interpreting}, using the official implementation provided by the authors. 
For all three datasets, we report performance using SpLiCE with an external vocabulary of 10,000 LAION-based concepts and an $\ell_1$ regularization weight of 0.25, following their best reported configuration. 
Despite relying on no external vocabulary, \ours consistently outperforms SpLiCE across all benchmarks highlighting the strength of sparse concept steering even in the absence of external lexical resources.

\begin{table}[H]
\centering
\caption{Zero-shot top-1 accuracy (\%) on \textbf{CIFAR-100}.}
\label{tab:cifar_vs_splice}
% \begin{adjustbox}{max width=\columnwidth}
\begin{tabular}{l cc}
\toprule
\textbf{Method} & ViT-B/32 & ViT-B/16 \\
\midrule
$\text{CLIP}_{\text{ZS}}$                     & \accinline{61.07}{0}           & \accinline{63.96}{0}           \\
$\text{SAE}_{\text{REC}}^{\text{A}}$          & \accinline{58.01}{\negd{–3.06}} & \accinline{64.05}{\posd{+0.09}} \\
$\text{SAE}_{\text{REC}}^{\text{F}}$          & \accinline{58.22}{\negd{–2.85}} & \accinline{63.42}{\negd{–0.54}} \\
$\text{SAE}_{\text{REC}}^{\text{F}+\gamma}$   & \accinline{62.69}{\posd{+1.62}} & \accinline{66.81}{\posd{+2.85}} \\
$\text{SpLiCE}$ \citep{bhalla2024interpreting} & \accinline{55.57}{\negd{–5.50}} & \accinline{58.29}{\negd{–5.67}} \\
\textbf{VS2 (ours)}                          & \accinline{\textbf{64.52}}{\posd{\textbf{+3.45}}} & \accinline{\textbf{68.08}}{\posd{\textbf{+4.12}}} \\
\bottomrule
\end{tabular}
% \end{adjustbox}
\end{table}

\section{Effect of Top-\(N\) Retrieved Neighbors}
\label{app:retrieved}
\begin{figure*}[t]
  \centering
  \includegraphics[width=0.95\textwidth]{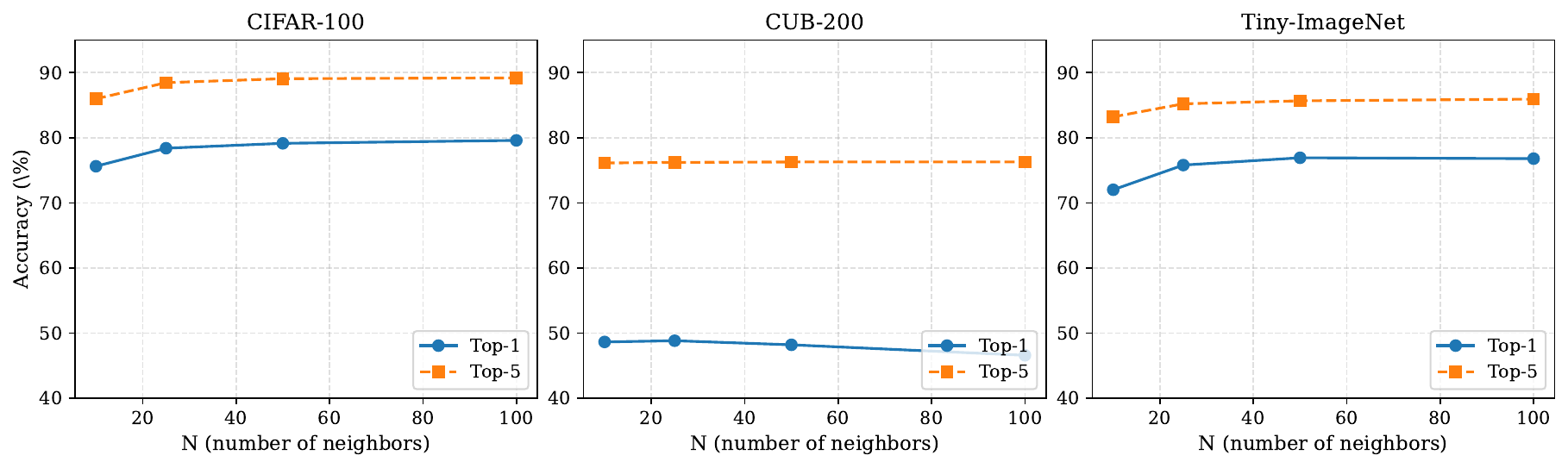}
  \caption{Top-1 and Top-5 accuracy as a function of number of retrieved neighbors \(N\), using ViT-B/16 (Patch-16) across datasets. Larger \(N\) improves general classification but may degrade performance in fine-grained settings.}
  \label{fig:accuracy_vs_n}
\end{figure*}

To assess the influence of retrieval size on performance, we conduct an ablation over the number of retrieved neighbors \(N\) used for latent aggregation. We use a fixed ViT-B/16 (Patch-16) backbone and vary \(N \in \{10, 25, 50, 100\}\) across three datasets: CIFAR-100, CUB-200, and Tiny-ImageNet. As shown in Figure~\ref{fig:accuracy_vs_n}, both top-1 and top-5 classification accuracy steadily increase with \(N\) on CIFAR-100 and Tiny-ImageNet, with diminishing returns beyond \(N=50\). In contrast, performance on CUB-200 remains largely flat or slightly degrades, suggesting that retrieving more neighbors in fine-grained datasets can introduce noise due to overly similar but semantically irrelevant examples. These results indicate that the utility of neighbor-based aggregation is dataset-dependent: general object recognition tasks may benefit from larger \(N\), while fine-grained classification may require more careful control of retrieval scope.

\section{Per-class Performance Analysis on CIFAR-100}
\label{app:performance}

In Table~\ref{tab:vs2_vs2pp_gainloss}, we present results for the classes most affected by steering CLIP ViT-B/32 with \ours{} and \oursplus. We highlight both positive and negative shifts in accuracy to identify the most impacted categories. Unlike the average performance reported in the main experiments, this analysis shows that the top-5 class-level gains are generally larger in magnitude than the corresponding losses. Although the accuracy drops for misclassified categories are smaller in magnitude, they are non-negligible. As future work, it would be valuable to explore when steering should be applied or withheld to further maximize overall performance gains.

\section{Prototype-Aware Sparse Steering Vectors}
\label{app:prototype}

Our core methods, \ours and its retrieval-augmented variant \oursplus, enhance zero-shot classification by steering CLIP embeddings along directions identified by a top-$k$ Sparse Autoencoder (SAE). These directions correspond to latent \emph{features} that, ideally, align with class-discriminative concepts. Steering in these directions upweights what the model has learned to be important during reconstruction. This raises a central hypothesis: \textbf{the reconstruction task itself is to some extent sufficient to uncover features that are also relevant for classification}. In other words, there is a meaningful overlap between features that are important for reconstructing the CLS token and those that are predictive for the downstream task. In this section, drawing inspiration from prototype theory~\citep{rosch1973natural}, we investigate whether incorporating prototype information during SAE training can better align the features important for reconstruction with those that are critical for downstream classification.

The limited improvements observed on fine-grained datasets like CUB-200 and Tiny-ImageNet suggest that the challenge lies not just in identifying sparse features, but in uncovering the \emph{correct} ones. This shifts the central question from “What are the most important sparse features to select?” to a deeper inquiry: “Can the reconstruction objective alone reliably capture features that are most useful for classification and if not, how can task-relevant information be effectively incorporated?”

\begin{table*}[t]
\centering
\caption{Benchmarking Visual Sparse Steering: Zero-Shot Accuracy (\%) with and Without External Data on CIFAR-100, CUB-200, and Tiny-ImageNet using ViT-B/32 and ViT-B/16.}
\label{tab:pass_acc_delta_inline}
\scriptsize
\setlength{\tabcolsep}{2.5pt}
\renewcommand{\arraystretch}{0.92}
\resizebox{\textwidth}{!}{
\begin{tabular}{l cc cc cc}
\toprule
& \multicolumn{2}{c}{\textbf{CIFAR-100}} 
& \multicolumn{2}{c}{\textbf{CUB-200}}
& \multicolumn{2}{c}{\textbf{Tiny-IN}} \\
\cmidrule(lr){2-3}\cmidrule(lr){4-5}\cmidrule(lr){6-7}
\textbf{Method}
  & ViT-B/32 & ViT-B/16
  & ViT-B/32 & ViT-B/16
  & ViT-B/32 & ViT-B/16 \\
\midrule
\multicolumn{7}{c}{\textbf{Zero-shot (no retrieval)}} \\[2pt]

$\text{CLIP}_{\text{ZS}}$ &
  \accinline{61.07}{0} &
  \accinline{63.96}{0} &
  \accinline{51.76}{0} &
  \accinline{55.06}{0} &
  \accinline{56.64}{0} &
  \accinline{61.08}{0} \\

$\text{SAE}_{\text{REC}}^{\text{A}}$ &
  \accinline{58.01}{\negd{--3.06}} &
  \accinline{64.05}{\posd{+0.09}} &
  \accinline{47.45}{\negd{-4.31}} &
  \accinline{51.81}{\negd{-3.25}} &
  \accinline{30.56}{\negd{--26.08}} &
  \accinline{52.96}{\negd{-8.12}} \\

$\text{SAE}_{\text{REC}}^{\text{F}}$ &
  \accinline{58.22}{\negd{--2.85}} &
  \accinline{63.42}{\negd{--0.54}} &
  \accinline{48.08}{\negd{-3.68}} &
  \accinline{51.43}{\negd{-3.63}} &
  \accinline{36.33}{\negd{--20.31}} &
  \accinline{54.84}{\negd{-6.24}} \\

$\text{SAE}_{\text{REC}}^{\text{F}+\gamma}$ &
  \accinline{62.69}{\posd{+1.62}} &
  \accinline{66.81}{\posd{+2.85}} &
  \accinline{49.41}{\negd{-2.35}} &
  \accinline{53.28}{\negd{-1.78}} &
  \accinline{39.49}{\negd{--17.15}} &
  \accinline{58.81}{\negd{-2.27}} \\

\textbf{VS2 (ours)} &
  \accinline{\textbf{64.52}}{\posd{\textbf{+3.45}}} &
  \accinline{\textbf{68.08}}{\posd{\textbf{+4.12}}} &
  \accinline{\textbf{52.69}}{\posd{\textbf{+0.93}}} &
  \accinline{\textbf{56.14}}{\posd{\textbf{+1.08}}} &
  \accinline{\textbf{58.14}}{\posd{\textbf{+1.50}}} &
  \accinline{\textbf{62.92}}{\posd{\textbf{+1.84}}} \\

\textbf{VS2 + PASS (ours)} &
  \accinline{\textbf{70.64}}{\posd{\textbf{+9.57}}} &
  \accinline{\textbf{68.23}}{\posd{\textbf{+4.27}}} &
  \accinline{\textbf{52.97}}{\posd{\textbf{+1.21}}} &
  \accinline{\textbf{56.63}}{\posd{\textbf{+1.57}}} &
  \accinline{\textbf{57.94}}{\posd{\textbf{+1.30}}} &
  \accinline{\textbf{62.98}}{\posd{\textbf{+1.90}}} \\
\bottomrule
\end{tabular}
}
\end{table*}

%--------------------------
\paragraph{\textbf{Oracle steering with known prototypes.}} Following prototype theory, we collect for every class the ten images that
ViT-B/32 CLIP classifies with the highest confidence; these serve as
\emph{oracle prototypes}.  
Given the true class label~$y$, we build a prototype steering vector by averaging their latent sparse features. \textbf{\ours using oracle prototypes can lift
CLIP to 97.5\% on CIFAR-100, 91.04\% on CUB-200, and 90.1\% on
Tiny-ImageNet}.  
This confirms the existence of discriminative sparse directions. In Appendix \ref{app:proto_orthogonality} we further examine the discriminative ability (measured by orthogonality) of these steering vectors. Generally, these prototype vectors have a low cosine similarity, yet a non-negligible
tail reveals strongly overlapping directions between visually or semantically close categories.

%--------------------------
\paragraph{\textbf{Prototype-aligned SAE (PASS).}}
Above, we constructed steering vectors using the pretrained SAE features of oracle prototypes. Now, we consider whether prototypes can be used to learn new, more informative SAE features. We assume during SAE training that we have access to class labels. Then, to the SAE loss we add a regularization term which encourages SAE features to be close to their class mean. That is, for a training sample~$i$ with latent sparse feature ~$\mathbf z_i$ and class mean
$\bar{\mathbf z}_{\text{class}(i)}$ we minimize  
\begin{equation}
\mathcal L = \mathcal L_{\text{recon}}
           + w_{\text{aux}} \;\bigl\lVert \mathbf z_i - \bar{\mathbf z}_{\text{class}(i)} \bigr\rVert_2^2,
\end{equation}
where $w_{\text{aux}}$ controls the strength of the prototype-alignment term relative to the reconstruction loss and is set to 0.8.
We refer to the resulting steering method as \textbf{PASS} (Prototype-Aligned Sparse Steering). Although PASS uses class labels during SAE training, it remains \emph{fully test-time unsupervised}. Empirically, in Table \ref{tab:pass_acc_delta_inline}, we observe that PASS outperforms \ours across all datasets, with particularly substantial gains on CIFAR-100. However, this improvement comes at the cost of requiring labels for each training sample during SAE training. Gains are modest on CUB-200 and Tiny-ImageNet, and we thus hypothesize 
that classes which share many features require richer or multi-prototype guidance which is an
intriguing avenue for future work.

\begin{table}[h]
\centering
\caption{\textbf{Trade-off between reconstruction fidelity and prototype alignment.} Increasing $w_{\text{aux}}$ improves classification accuracy but degrades FVU reconstruction.}
\label{tab:recon-align-tradeoff}
\small
\begin{tabular}{c c c}
\toprule
$w_{\text{aux}}$ & FVU $\downarrow$ & Accuracy (\%) $\uparrow$ \\
\midrule
0.1 & 0.3437 & 68.80 \\
0.5 & 0.4887 & 70.40 \\
1.0 & 0.5393 & 70.61 \\
2.0 & 0.5702 & 70.76 \\
\bottomrule
\end{tabular}
\end{table}

\paragraph{\textbf{Reconstruction vs. Alignment Trade-off}}
\label{sec:recon-align-tradeoff}

Sparse Autoencoders (SAEs) trained for reconstruction can also be optimized to align their latent features with class-level prototypes. However, this introduces a trade-off between two competing objectives: fidelity of reconstruction and discriminative alignment.

To investigate this trade-off, we introduce a weighting coefficient $w_{\text{aux}}$ that controls the strength of prototype alignment relative to the reconstruction loss. As $w_{\text{aux}}$ increases, alignment is encouraged more strongly. Table~\ref{tab:recon-align-tradeoff} reports the resulting changes in reconstruction loss (measured by FVU) and top-1 classification accuracy on CIFAR-100 using ViT-B/16, with 128 sparse latents and expansion factor 4.

We observe that increasing $w_{\text{aux}}$ consistently improves classification performance, up to a point, even though it introduces more reconstruction error. This aligns with our hypothesis: while exact input reconstruction encourages general feature coverage, alignment with class prototypes promotes discriminative feature extraction. These results highlight the flexibility of SAE-based steering to balance interpretability and performance depending on downstream objectives.

\section{Are exemplar-derived directions really distinct?}
\label{app:proto_orthogonality}

A desirable property of class–specific steering vectors is \emph{orthogonality}:
pushing an embedding toward class \emph{A} should not simultaneously raise its
score for class \emph{B}.  
Using the oracle prototypes as described in Appendix \ref{app:prototype}, we compute for
every CIFAR-100 class a prototype steering vector
and measure the pair-wise cosine similarity at layer 11
of the ViT-B/32 encoder.  
Most pairs have low similarity (mean=0.23), yet a non-negligible
tail reveals strongly overlapping directions.  
Table \ref{tab:proto_overlap} lists the ten highest-overlap pairs.

\begin{table}[h]
\centering
\caption{\textbf{Top-10 most overlapping prototype steering directions on CIFAR-100}.
High cosine similarity indicates that the two classes share visual attributes
that the SAE encodes along nearly the same latent axis.}
\label{tab:proto_overlap}
\small
\setlength{\tabcolsep}{7pt}
\begin{tabular}{@{}c l l S[table-format=1.2]@{}}
\toprule
Rank & \multicolumn{1}{c}{Class 1} & \multicolumn{1}{c}{Class 2} & {Cosine $\uparrow$} \\
\midrule
1  & beetle      & cockroach    & 0.91 \\
2  & mouse       & shrew        & 0.89 \\
3  & dolphin     & shark        & 0.84 \\
4  & otter       & seal         & 0.84 \\
5  & dolphin     & whale        & 0.84 \\
6  & possum      & raccoon      & 0.84 \\
7  & snake       & worm         & 0.83 \\
8  & oak tree    & willow tree  & 0.83 \\
9  & ray         & shark        & 0.81 \\
10 & bowl        & cup          & 0.80 \\
\bottomrule
\end{tabular}
\end{table}

\begin{table*}[t]
\centering
\caption{Training-time compute and parameter cost per sample for adaptation strategies on the CLIP ViT-B/32 vision encoder.}
\label{tab:compute-train}
\small
\setlength{\tabcolsep}{5pt}
\renewcommand{\arraystretch}{0.95}
\begin{tabular}{lccc}
\toprule
\textbf{Method} & \textbf{Train FLOPs / sample} & \textbf{GFLOPs} & \textbf{Trainable Params} \\
\midrule
SAE (cached activations)             & $1.42{\times}10^{7}$  & 0.014  & 4.72M \\
SAE (non-cached; incl.\ extraction)  & $8.74{\times}10^{9}$  & 8.744  & 4.72M \\
LoRA (rank = 16)                     & $2.64{\times}10^{10}$ & 26.365 & 0.59M \\
Full fine-tuning                     & $2.62{\times}10^{10}$ & 26.188 & 87.46M \\
\bottomrule
\end{tabular}
\end{table*}

These high-overlap pairs are \emph{semantically plausible} confusions
(e.g.\ beetle vs.\ cockroach or dolphin vs.\ whale),
confirming that exemplar steering directions tend to align for
visually or taxonomically proximate classes.  
In downstream applications, a simple orthogonalization step may help reduce feature overlap between sparse directions. Investigating principled ways to encourage orthogonality during SAE training is a promising direction for future work. 

\begin{table*}[t]
\centering
\caption{\textbf{Top-5 class gains and losses on CIFAR-100.} Green = absolute gain; Red = absolute loss relative to ZS baseline.}
\label{tab:vs2_vs2pp_gainloss}

\begin{tabular*}{\textwidth}{@{\extracolsep{\fill}}cc@{}}

\subcaptionbox{\textbf{VS2 – Top-5 Gains}}[0.5\textwidth]{%
\small
\begin{adjustbox}{max width=\linewidth}
\begin{tabular}{p{1.8cm} S p{1.8cm} }
\toprule
Class & {ZS} & \textbf{VS2 \gain{($\Delta$)}}  \\
\midrule
tractor         & 0.55 & 0.80 \gain{(+0.25)}  \\
forest          & 0.35 & 0.58 \gain{(+0.23)}  \\
man             & 0.53 & 0.74 \gain{(+0.21)}  \\
bus             & 0.51 & 0.68 \gain{(+0.17)} \\
snake           & 0.61 & 0.76 \gain{(+0.15)}  \\
\bottomrule
\end{tabular}
\end{adjustbox}}

&
% VS2 Losses
\subcaptionbox{\textbf{VS2 – Top-5 Losses}}[0.5\textwidth]{%
\small
\begin{adjustbox}{max width=\linewidth}
\begin{tabular}{p{1.8cm} S p{1.8cm} }
\toprule
Class & {ZS} & \textbf{VS2 \loss{($\Delta$)}}  \\
\midrule
aquarium\_fish & 0.82 & 0.61 \loss{(-0.21)}  \\
beetle         & 0.64 & 0.49 \loss{(-0.15)}  \\
sweet\_pepper  & 0.70 & 0.58 \loss{(-0.12)}  \\
tulip          & 0.78 & 0.71 \loss{(-0.07)}  \\
maple\_tree    & 0.57 & 0.50 \loss{(-0.07)}  \\
\bottomrule
\end{tabular}
\end{adjustbox}}

\\[2ex]

\subcaptionbox{\textbf{VS2++ – Top-5 Gains}}[0.5\textwidth]{%
\small
\begin{adjustbox}{max width=\linewidth}
\begin{tabular}{p{1.8cm} S p{1.8cm} }
\toprule
Class & {ZS} & \textbf{VS2++ \gain{($\Delta$)}} \\
\midrule
spider          & 0.48 & 0.91 \gain{(+0.43)}  \\
caterpillar     & 0.27 & 0.68 \gain{(+0.41)}  \\
possum          & 0.24 & 0.65 \gain{(+0.41)}  \\
tractor         & 0.55 & 0.96 \gain{(+0.41)}  \\
tiger           & 0.45 & 0.84 \gain{(+0.39)}  \\
\bottomrule
\end{tabular}
\end{adjustbox}}

&
% VS2++ Losses
\subcaptionbox{\textbf{VS2++ – Top-5 Losses}}[0.5\textwidth]{%
\small
\begin{adjustbox}{max width=\linewidth}
\begin{tabular}{p{1.8cm} S p{1.8cm} }
\toprule
Class & {ZS} & \textbf{VS2++ \loss{($\Delta$)}}  \\
\midrule
girl           & 0.72 & 0.65 \loss{(-0.07)}  \\
maple\_tree    & 0.57 & 0.51 \loss{(-0.06)}  \\
porcupine      & 0.18 & 0.13 \loss{(-0.05)} \\
ray            & 0.06 & 0.02 \loss{(-0.04)}  \\
mouse          & 0.18 & 0.15 \loss{(-0.03)}  \\
\bottomrule
\end{tabular}
\end{adjustbox}}

\end{tabular*}
\end{table*}

\section{Computational Overhead: Additional Details}
\label{app:compute}
\paragraph{Training-time FLOPs.}
To quantify the cost of training the SAE versus alternative adaptation methods, Table~\ref{tab:compute-train}
reports per-sample training FLOPs (forward + backward) and the number of trainable parameters for the
CLIP ViT-B/32 vision encoder. Even when activation extraction is included, SAE training remains more than $3\times$ cheaper than LoRA or full
fine-tuning. With cached activations, as in our VS2 setup, the cost drops to just 0.014~GFLOPs per sample.

\paragraph{TPT FLOP accounting.}
For completeness, we outline how the $16.6\times$ test-time overhead factor for TPT~\citep{tpt} is obtained.
A single CLIP ViT-B/32 forward pass on a $224 \times 224$ image with 100 text prompts costs $8.7$~GFLOPs for the
vision encoder and $581.8$~GFLOPs for the text encoder, for a total of approximately $0.59$~TFLOPs. In our
implementation, TPT performs $63$ augmented views per image and runs $4$ optimization steps; each step requires a
full CLIP forward pass over all views and a backward pass through the text encoder. This yields about
$9.217$~TFLOPs of adaptation compute per test image, followed by one final forward-only evaluation of
$0.591$~TFLOPs, for a total of roughly $9.807$~TFLOPs. Dividing by the $0.59$~TFLOPs of plain CLIP gives the
reported $\approx 16.6\times$ test-time compute overhead.

\begin{table}[t]
% \label{tab:cosine}
\centering
\caption{\textbf{Cosine similarity ($\uparrow$) between steering vectors learned by different SAE capacities.}
Rows/columns are ordered by expansion factor and sparsity (\emph{e} × expansion, $k$ active).}
\label{tab:sae-sim-no-clip}
\small
\setlength{\tabcolsep}{4pt}
\renewcommand{\arraystretch}{1.12}
\begin{adjustbox}{max width=\textheight,center}
\begin{tabular}{l*{13}{c}}
\toprule
\textbf{SAE cfg.} &
\rotatebox{60}{16×/64} &
\rotatebox{60}{16×/512} &
\rotatebox{60}{16×/128} &
\rotatebox{60}{16×/256} &
\rotatebox{60}{8×/128} &
\rotatebox{60}{8×/512} &
\rotatebox{60}{8×/256} &
\rotatebox{60}{4×/256} &
\rotatebox{60}{4×/512} &
\rotatebox{60}{4×/128} &
\rotatebox{60}{10×/128} &
\rotatebox{60}{4×/64} &
\rotatebox{60}{8×/64} \\
\midrule
16×/64   & \textbf{1.000} & 0.497 & 0.577 & 0.705 & 0.726 & 0.635 & 0.680 & 0.640 & 0.591 & 0.681 & 0.536 & 0.701 & 0.716 \\[2pt]
16×/512  & 0.497 & \textbf{1.000} & 0.161 & 0.399 & 0.358 & 0.817 & 0.634 & 0.618 & 0.611 & 0.584 & 0.136 & 0.316 & 0.282 \\[2pt]
16×/128  & 0.577 & 0.161 & \textbf{1.000} & 0.900 & 0.865 & 0.273 & 0.298 & 0.277 & 0.265 & 0.303 & 0.952 & 0.887 & 0.678 \\[2pt]
16×/256  & 0.705 & 0.399 & 0.900 & \textbf{1.000} & 0.956 & 0.527 & 0.572 & 0.495 & 0.410 & 0.542 & 0.920 & 0.936 & 0.744 \\[2pt]
8×/128   & 0.726 & 0.358 & 0.865 & 0.956 & \textbf{1.000} & 0.543 & 0.650 & 0.575 & 0.483 & 0.634 & 0.896 & 0.959 & 0.829 \\[2pt]
8×/512   & 0.635 & 0.817 & 0.273 & 0.527 & 0.543 & \textbf{1.000} & 0.861 & 0.811 & 0.771 & 0.796 & 0.243 & 0.483 & 0.506 \\[2pt]
8×/256   & 0.680 & 0.634 & 0.298 & 0.572 & 0.650 & 0.861 & \textbf{1.000} & 0.889 & 0.771 & 0.912 & 0.289 & 0.557 & 0.624 \\[2pt]
4×/256   & 0.640 & 0.618 & 0.277 & 0.495 & 0.575 & 0.811 & 0.889 & \textbf{1.000} & 0.873 & 0.941 & 0.240 & 0.532 & 0.600 \\[2pt]
4×/512   & 0.591 & 0.611 & 0.265 & 0.410 & 0.483 & 0.771 & 0.771 & 0.873 & \textbf{1.000} & 0.821 & 0.191 & 0.461 & 0.549 \\[2pt]
4×/128   & 0.681 & 0.584 & 0.303 & 0.542 & 0.634 & 0.796 & 0.912 & 0.941 & 0.821 & \textbf{1.000} & 0.284 & 0.594 & 0.655 \\[2pt]
10×/128  & 0.536 & 0.136 & 0.952 & 0.920 & 0.896 & 0.243 & 0.289 & 0.240 & 0.191 & 0.284 & \textbf{1.000} & 0.910 & 0.678 \\[2pt]
4×/64    & 0.701 & 0.316 & 0.887 & 0.936 & 0.959 & 0.483 & 0.557 & 0.532 & 0.461 & 0.594 & 0.910 & \textbf{1.000} & 0.814 \\[2pt]
8×/64    & 0.716 & 0.282 & 0.678 & 0.744 & 0.829 & 0.506 & 0.624 & 0.600 & 0.549 & 0.655 & 0.678 & 0.814 & \textbf{1.000} \\
\bottomrule
\end{tabular}
\end{adjustbox}
\end{table}

\end{document}